\documentclass[runningheads]{llncs}

\usepackage[table,xcdraw]{xcolor}
\usepackage{graphicx}

\usepackage{xspace}
\usepackage{appendix}

\usepackage{caption,subcaption}
\usepackage{amsmath}
\usepackage{amsfonts}

\usepackage{tikz}
\usetikzlibrary{positioning}

\usepackage{circuitikz}
\graphicspath{ {./TIKS_images/} }

\usepackage{algorithm}
\usepackage{algorithmic}

\usepackage[labelformat=simple]{subcaption}

\usepackage[colorlinks=true, allcolors=blue]{hyperref} 

\usepackage{xspace}

\usepackage{color}

\usepackage{booktabs}
\usepackage{threeparttable}
\usepackage{cellspace, makecell, multirow}

\usepackage[normalem]{ulem} 

\definecolor{cgrey}{gray}{0.65}
\definecolor{ccgrey}{gray}{0.85}
\definecolor{greenhtml}{HTML}{C6EFCE}

\newcolumntype{g}{>{\columncolor{greenhtml}}c}
\newcolumntype{b}{>{\color{cgrey}}c}

\newcommand\cmark[1][]{%
    \tikz\fill[scale=0.35](0,.35) -- (.25,0) -- (1,.7) -- (.25,.15) -- cycle;}%
\newcommand\crossmark[1][]{%
  \tikz[scale=0.25,#1]{
    \fill(0,0)--(0.1,0) .. controls (0.5,0.4) .. (1,0.7)--(0.9,0.7) ..  controls (0.5,0.5) ..(0,0.1) --cycle;
    \fill(1,0.1)--(0.9,0.1) .. controls (0.5,0.3) .. (0,0.7)--(0.1,0.7) .. controls (0.5,0.4) ..(1,0.2) --cycle;
  }%
}

\newcommand{\SOE}{\emph{SOE}\xspace}
\newcommand{\LastLayer}{\emph{Last Hidden Layer}\xspace}
\newcommand{\NeuronWiggle}{\emph{Neuron Wiggle}\xspace}
\newcommand{\Freeze}{\emph{Freeze}\xspace}
\newcommand{\ExhaustiveS}{\emph{Exhaustive Search}\xspace}

\newcommand{\dotp}[2]{\langle #1, #2 \rangle}
\newcommand{\norm}[1]{\| #1 \|}

\begin{document}
\title{Polynomial Time Cryptanalytic Extraction of Neural Network Models}

%

\author{
Isaac A. Canales-Mart\'inez\inst{2} \and
Jorge Chavez-Saab\inst{2} \and
Anna Hambitzer\inst{2} \and
Francisco Rodr\'iguez-Henr\'iquez\inst{2} \and 
Nitin Satpute\inst{2} \and
Adi Shamir\inst{1} 
}

%
\authorrunning{ }
%
\institute{
Weizmann Institute
\email{adi.shamir@weizmann.ac.il}\\
\and
Cryptography Research Center, Technology Innovation Institute\\
\email{\{isaac.canales,anna.hambitzer,jorge.saab, francisco.rodriguez,nitin.satpute\}@tii.ae}}
%
\maketitle              
\begin{abstract}
Billions of dollars and countless GPU hours are currently spent on training Deep Neural Networks (DNNs) for a variety of tasks. Thus, it is essential to determine the difficulty of extracting all the parameters of such neural networks when given access to their black-box implementations. Many versions of this problem have been studied over the last 30 years, and the best current attack on ReLU-based deep neural networks was presented at Crypto'20 by Carlini, Jagielski, and Mironov. It resembles a differential chosen plaintext attack on a cryptosystem, which has a secret key embedded in its black-box implementation and requires a polynomial number of queries but an exponential amount of time (as a function of the number of neurons).  

In this paper, we improve this attack by developing several new techniques that enable us to extract with arbitrarily high precision all the real-valued parameters of a ReLU-based DNN using a polynomial number of queries \emph{and} a polynomial amount of time. We demonstrate its practical efficiency by applying it to a full-sized neural network for classifying the CIFAR10 dataset, which has 3072 inputs, 8 hidden layers with 256 neurons each, and about $1.2$ million neuronal parameters. An attack following the approach by Carlini et al.\ requires an exhaustive search over $2^{256}$ possibilities. 
Our attack replaces this with our new techniques, which require only 30 minutes on a 256-core computer.

\keywords{ReLU-Based Deep Neural Networks \and Neural Network Extraction \and Polynomial Query and Polynomial Time Differential Attack.}
\end{abstract}

\section{Introduction}\label{sec:intro}

In this paper, we consider the problem of extracting all the weights and biases of a Deep Neural Network (DNN), which is queried as a black box to obtain the outputs corresponding to carefully chosen inputs. This is a long-standing open problem that was first addressed by cryptographers and mathematicians in the early nineties of the last century~\cite{BlumR93,Fefferman1994}, and then was followed up one decade later by an adversarial reverse-engineering attack presented by Lowd and Meck in~\cite{LowdM05}. More recently, since around 2016, this problem has enjoyed a steady stream of new ideas by research groups from both industry and academia~\cite{Oliynyk_2023,jagielski2020high,RolnickK20,BatinaBJP19,tramer2016stealing}, culminating with an algorithm that Carlini, Jagielski and Mironov presented at Crypto 2020~\cite{carlini2020cryptanalytic}, which in the general case, requires a polynomial number of queries but an exponential amount of time. Due to this time complexity, their algorithm was practically demonstrated only on some toy examples in which the layers of the showcased networks (beyond the first hidden layer, which is usually easy to extract) contain at most a few tens of neurons.

The main obstacle that made the Carlini et al.\ algorithm exponential in time was their inability to determine one bit of information (namely, a $\pm$ sign) about each neuron in the DNN. This hindrance forced them to use exhaustive search over all the possible sign combinations in each layer, which is prohibitively expensive for any realistic network whose width is more than a few tens of neurons. Our main contribution in this paper is to show how to efficiently find these signs by a new chosen input attack, which we call {\it neuron wiggling}, as well as two other methods that target the first two hidden layers and the last one. Our techniques resemble a combination of first-order and second-order differential cryptanalysis, in which we use a chosen input attack to slightly change the inputs to the DNN in a small number of carefully chosen directions and observe both the slope and the curvature produced in the output of the network as a result of these input changes. However, it differs from standard differential attacks on digital cryptosystems by using real-valued inputs and outputs rather than bit strings.

We demonstrate the practical applicability of our algorithm by performing a sign-recovery attack on a DNN trained to recognize the standard CIFAR10 classes. This DNN has 3072 inputs, 8 hidden layers, 256 neurons per hidden layer, and about $1.2$ million weights. Our techniques replaced the exhaustive search over $2^{256}$ possible combinations of neuronal signs carried out by Carlini et al.\ in each layer by a new algorithm which requires only 32 minutes on a $256$-core computer to find all the $8 \times 256$ neuronal signs in the $8$ layers of the network.

Our model of deep neural networks is essentially the same as the one used by Carlini et al.. We assume that the network is fully connected (with no skip connections), with $r$ hidden layers of varying width in which each neuron consists of an affine mapping followed by Rectified Linear Units (ReLU) activation functions. For the sake of simplicity in the analysis of our algorithm, we assume that all the network weights are real numbers, that arithmetic operations over real numbers (both by the network and by the attacker) are carried out with infinite precision in unit time, and that the attacker can perform arbitrarily small changes in the inputs of the network and observe their corresponding outputs with arbitrarily high precision. In practice, it was sufficient to use standard 64-bit floating point arithmetic to extract the full-sized CIFAR10 network, but higher precision may be required to attack considerably deeper networks due to the possible accumulation of rounding errors throughout our computations. However, since our algorithm uses only polynomially many arithmetic operations, we expect the asymptotic number of bits of precision to also grow only polynomially with the size of the network.

In our attack (as well as the one by Carlini et al.), it is essential to assume that the activation function is ReLU-like since we concentrate on the behavior of the network in the vicinity of {\it critical points}, which are defined as inputs in whose tiny vicinity exactly one ReLU input in one of the layers changes sign. Regardless of the complexity of the network, we expect the network's output to abruptly change its behavior as a result of this ReLU transition, which makes this event noticeable when we observe the network's outputs. Such abrupt changes are also visible when we apply a smooth softmax function to the logits, and thus, our attack can use outputs which are either the logits produced by the last linear layer or their softmax values (which translate the logits into a probability distribution over the possible classes). We can generalize the attack to any other type of activation function that looks like a piecewise linear function (with possibly more than one sharp bend), such as a leaky ReLU, but not to smooth activation functions such as a sigmoid which have no critical points. Our attack can be applied without modifications to convolutional networks since they can be described as a special form of a fully connected network.

While our analysis and experiments support our claim of efficiency in the extraction of the weights of large networks, there are several important caveats. First of all, fully connected networks have certain symmetries which make it impossible to extract the exact form of the original network, but only a functionally equivalent form. For example, we can reorder the neurons of any layer and adjust the weights accordingly, we can merge two neurons that have exactly the same weights and biases, we can eliminate any neuron whose output is multiplied by a weight of zero by all the next layer neurons, or we can multiply all the weights (including the bias) of a particular neuron by some positive constant $c$ while multiplying all the weights which multiply the output of that neuron in the next layer neurons by $c^{-1}$; all these changes do not change the input/output behavior of the network, and thus they are invisible to any black-box attack. In addition, there could be some unlucky events, which we do not expect to encounter but which will cause our attack to fail. For example, there could be neurons whose values before the ReLU almost never change sign, and thus our bounded number of queries will fail to detect any critical point for them.\footnote{We refer to these neurons as almost always-on/off neurons and note that the almost perfectly linear behavior of such neurons will be simply absorbed by the linear mappings in the next layer, which will result in a very close approximation of the original neural network. In practice, in our networks, always-on neurons were only encountered immediately after random initialization. After training, however, this kind of neuron was never encountered, but we did find instances of always-off neurons. 
} Finally, during the attack, we may fail to solve some systems of linear equations if their determinant is exactly zero, but even the slightest change in the real-valued entries in such a matrix should eliminate the problem. We thus have to qualify our assertion that our attack always succeeds, in the same way that most cryptanalytic attacks on cryptosystems are in fact heuristics that are not guaranteed to succeed in some particularly unlucky situations.

It is important to note that when the attacker is only given an (adversarially chosen) collection of known inputs along with their corresponding outputs, then a famous result of Blum and Rivest back in 1993~\cite{BlumR93} shows that just to decide whether there exists a 2 layer 3 neuron DNN on $n$ inputs which is consistent with the provided pairs is NP-complete. However, in our case, the attacker can apply a chosen input attack, and thus he is not likely to suffer from this particular complexity barrier.

\subsection{Our Contributions}

Building on the work of~\cite{carlini2020cryptanalytic}, 
we present a black-box attack that permits the recovery of a functionally equivalent model of a deep neural network that uses ReLU activation functions. 
In contrast with the approach presented in~\cite{carlini2020cryptanalytic}, our attack has polynomial time complexity in terms of the number of neurons in the DNN, and can thus be applied (in principle) to arbitrarily deep and wide neural networks. In addition, we can easily deal with many types of expanding neural networks, where~\cite{carlini2020cryptanalytic} struggles.

To make this possible, we develop three new sign recovery techniques. The first one, called \emph{SOE}, is based on solving a system of linear equations derived from observed first derivative values. The \emph{Neuron Wiggle} method applies differences in the input to maximally change the value of the neuron of interest. These changes propagate to the output and a statistical analysis of the output variations allows recovery of the sign. Finally, the \emph{Last Hidden Layer} technique is applicable only to the last layer in the network and employs second-order derivatives to construct a different system of linear equations whose solution simultaneously yields the signs of all the neurons in this layer.

We showcase our findings with a practical sign-recovery attack against a 3072-input network with 8 hidden layers and $d=256$ neurons per hidden layer for classifying the CIFAR10 dataset.

\subsection{Overview of our Attack}
\label{sec:Introduction_OverviewOfAttack}

Our attack follows the same strategy as the one presented in \cite{carlini2020cryptanalytic} by Carlini et al. Namely, we recover the parameters (i.e., weights and bias of each neuron) for the first hidden layer and ``peel off'' that layer. Thus, the attack now reduces to extracting the parameters of a DNN with one less hidden layer. We then recover the parameters for the second layer, peel it off, and continue in this fashion until the last hidden layer. Finally, the parameters for the output layer\footnote{By the output layer, we mean the very last layer of the DNN. We assume that this layer is linear in the sense that it doesn't contain ReLUs. The output layer is not counted as a hidden layer of the DNN.} are obtained. After peeling off the first hidden layer, however, we no longer have full control of the input to the second layer since the first layer is not an invertible mapping (e.g., no negative numbers can be output by its ReLUs). This lack of control over the input to the layer is also true for all subsequent hidden layers.

Recovering the parameters of each hidden layer is done in two steps. First, for each neuron, we recover some multiple of its weights and bias. The sign of this unknown multiplicative factor is called the \emph{sign for the neuron}. In the second step, we find the signs for all the neurons in the current layer; we need them to find the actual mapping represented by this layer in order to peel it off and proceed to the next layer. Recovering signs is essential because multiplication by a positive constant is a symmetry of the network, but multiplication by a negative constant is not a symmetry due to the nonlinear behavior of the ReLU. 

To find the multiple of the parameters, we use the same techniques as in \cite[Sections 4.2 and 4.3.2]{carlini2020cryptanalytic}. To recover the sign for the neurons in the general case, Carlini et al.\ had to use exhaustive search over all the sign combinations in all the neurons in the current layer, which required an exponential amount of time. In this work, we introduce three new polynomial-time techniques.

\textbf{System of equations (\SOE)}. As pointed out in~\cite{carlini2020cryptanalytic}, in networks that are contractive enough, we essentially have full control over the inputs to any of the layers, and the signs can be easily recovered one by one through a method they propose, which we refer to as \Freeze. Here, we present an alternative method called \SOE, which recovers the signs for all neurons simultaneously by solving a system of equations. Not only is it more efficient in terms of oracle queries, but also in time complexity since it solves a single system of equations whereas \Freeze would need to solve one system of equations per neuron. Both \Freeze and \SOE are significantly simpler and more efficient than the methods we describe next but are heavily limited by the contraction requirement and typically are only applicable to the first few layers.

\textbf{\NeuronWiggle}. Practically, most neural networks do not have contractive-enough hidden layers. The attack in \cite{carlini2020cryptanalytic} uses, in this case, a more general technique that determines the signs by exhaustive search while requiring a polynomial number of oracle queries. As our main result, we present the \NeuronWiggle technique, which is polynomial in both, time and queries to the oracle. This is the method of choice for most of the layers in both, contractive or expansive networks. In this method, we choose a \emph{wiggle} (i.e., a small change in a carefully chosen direction) in the input to the network that makes the input to the ReLU for some targeted neurons experience a large change, while all the other neurons in the network are expected to experience much smaller changes. This makes it possible to recover the sign for the targeted neuron, since if the input to that neuron's ReLU is negative, the effect of the large change will be blocked, while if the input to the ReLU is positive, this change will propagate through this ReLU and the subsequent layers, and eventually cause a change at the output. By choosing a critical point and wiggling in two opposite directions, we can detect the direction where the larger output change is present, thereby recovering the sign. However, some critical points may lead to a wrong decision on the sign, and thus, we have to repeat it at multiple unrelated inputs to gather reliable statistical evidence for which wiggles are blocked and which ones go through the ReLU. It is important to note that even without knowing the neuron's sign, the attacker can easily aggregate all the experiments that are on the same active/inactive side of the ReLU, and then he can use the statistical difference between these two clusters to determine which cluster is on the active and which cluster is on the inactive side of the ReLU. 

In our practical attacks, accumulating statistical evidence from 200 input points 
was sufficient to recover the correct sign for each neuron, except possibly in the last hidden layer. The main problem in the last hidden layer is that there is no randomization present in the function computing the output. 
That means if the weight that connects a particular neuron to the output is considerably smaller than the others, almost no amount of wiggling of the corresponding neuron will get through to the output. 
We thus used the neuron wiggle technique to successfully attack all but the first and last hidden layers of a CIFAR10 network with $256$ neurons per layer.

\textbf{\LastLayer}. Here, we develop a specialized technique to reliably deal with the last layer, which exploits exactly the same property that made the neuron wiggling technique less reliable for this layer (namely, the fact that there is a fixed linear mapping that maps the outputs of these neurons to the final output). This allows us to construct arbitrarily many linear equations in the coefficients of this fixed output function by exploring multiple unrelated inputs to the DNN, and their unique solution simultaneously yields the unknown signs for all the last layer neurons. This method has lower time and query complexity than Neuron Wiggle, but its applicability is limited to the last layer.

\subsection*{Organization}

The remainder of this paper is organized as follows. In~\autoref{sec:related}, we present a brief summary of the state of the art on DNN parameter extraction in the black-box model. In~\autoref{sec:prelim}, we present several basic definitions, and assumptions, and state the problem to solve. We complete that section by giving an overview of the approach by Carlini et al.\ before presenting our own sign-recovery techniques in \autoref{sec:sign_recovery_techniques}. All the practical attacks carried out in this work are presented in~\autoref{sec:experiments}. The concluding remarks are drawn in~\autoref{sec:conc}.

\section{Related Work} \label{sec:related}


Model extraction in the context of DNNs aims to obtain its architecture, the weights, and biases associated with each neuron in the network and, occasionally, the network's training hyperparameters. Although early work can be traced back to Fefferman~\cite{Fefferman1994} 
(who proved in 1994 that perfect knowledge of the output of a sigmoid-based network 
uniquely specifies its architecture and neurons’ weights, and also proved that two neural networks with the same input-output map are isomorphic up to trivial equivalences),
followed by Lowd and Meek in 2005~\cite{LowdM05}, the most important body of literature on this topic has been published since 2016, when Tram{\`e}r et al.\ studied the problem of functional equivalence for the case of multi-class logistic regression and the multi-layer perceptron. Since then, many attacks have been presented considering different attack scenarios and security models  \cite{abs-2105,carlini2020cryptanalytic,MilliSDH19,Reith0T19,PapernotMGJCS17,LowdM05} (see also~\cite{Oliynyk_2023} for a comprehensive survey).

As discussed in \autoref{sec:intro}, in this paper, we are primarily interested in a security model where the only interaction the attacker has with the DNN is submitting queries to it and observing the corresponding outputs (e.g., accessing it as a web service). 
Therefore, the attacker does not have access to the software or the hardware where the network has been deployed, which rules out side-channel and/or fault injection attacks as the ones reported in~\cite{joud2022practical,BatinaBJP19}. Furthermore, the main goal of our attack is to recover a functionally equivalent model of the network. Such precision is normally out of reach for the hardware attacks described in~\cite{joud2022practical}.

The state-of-the-art black-box model extraction has been traditionally centered around DNNs with ReLU activation functions. In the beginning, it was thought that the most effective way of extracting the model of the networks was through oracle calls that revealed information about the gradients. In~\cite{MilliSDH19}, this approach was theoretically analyzed and implemented to attack networks with one hidden layer. An improvement of this work followed quickly afterward in~\cite{jagielski2020high}, offering significant improvements to the functional equivalence of the extracted model but still constrained to attacks dealing with relatively modest one-layer DNNs and relying on the unrealistic help of an oracle that leaked the gradients.

Almost simultaneously, Rolnick and K\"ording presented in~\cite{RolnickK20} an approach that, by only observing the output of the network, was theoretically capable of extracting the parameters of deeper networks. However, in practice, the algorithm in~\cite{RolnickK20} only worked for two-layer networks. Meanwhile, a similar strategy was theoretically analyzed in~\cite{abs-2105}. A follow-up work by Carlini et al. was presented in~\cite{carlini2020cryptanalytic}. In this paper, the authors presented important technical improvements and devised novel techniques that allowed them to obtain a much higher precision in the neuron's parameters extraction while using remarkably fewer oracle calls than the previous methods reported in~\cite{RolnickK20,jagielski2020high} (cf. \cite[Table 1]{carlini2020cryptanalytic}). However impressive the high precision results obtained by the authors, they only managed to deal with neural networks of no more than three hidden layers and with a relatively modest number of a few tens of neurons beyond the first layer. 
The main reason the approach presented in~\cite{carlini2020cryptanalytic} could not scale up well for larger and deeper networks was that the weights of the neurons could only be obtained up to a constant of unknown sign. Indeed, finding the sign for the neurons had a prohibitively exponential cost in time and has remained until now as one of the two main obstacles towards extracting deeper and larger neural networks. The second obstacle, also acknowledged by the authors of~\cite{carlini2020cryptanalytic}, is that of dealing with so-called expansive neural networks, i.e., networks where the number of neurons in a given inner layer is larger than the number of inputs to that layer.

\section{Preliminaries}
\label{sec:prelim}

\subsection{Basic Definitions and Notation}

Informally, a neural network is a collection of connected nodes called \emph{neurons}. Neurons are arranged in \emph{layers} and are connected to those in the previous and the next layer. Every neuron has a \emph{weight} associated with each incoming connection and a \emph{bias}. We present next several important formal definitions by closely following the definitions and notation given in~\cite{carlini2020cryptanalytic}.

\begin{definition}
\label{def:DNN}
An \emph{$r$-deep neural network} is a function $f: \mathcal{X} \rightarrow \mathcal{Y}$ composed of alternating linear \emph{layers} $f_i$ and a non-linear \emph{activation function} $\sigma$ acting component-wise:
\[ f = f_{r+1} \circ \sigma \circ \cdots \circ \sigma \circ f_{2} \circ \sigma \circ f_{1}. \]
\end{definition}

We focus our study on Deep Neural Networks (DNN) over the real numbers. Then, $\mathcal{X} = \mathbb{R}^{d_0}$ and $\mathcal{Y} = \mathbb{R}^{d_{r+1}}$, where $d_0$ and $d_{r+1}$ are positive integers. As in~\cite{carlini2020cryptanalytic}, we only consider neural networks using the ReLU activation function $\sigma: x \mapsto \max(x, 0)$.

\begin{definition}
The $i$-th \emph{fully connected layer} of a neural network is a function $f_i: \mathbb{R}^{d_{i-1}} \rightarrow \mathbb{R}^{d_i}$ given by the affine transformation
\[ f_i(x) = A^{(i)}x + b^{(i)}, \]
where $A^{(i)} \in \mathbb{R}^{d_i \times d_{i-1}}$ is the \emph{weight matrix}, $b^{(i)} \in \mathbb{R}^{d_i}$ is the \emph{bias vector} of the $i$-th \emph{layer} and $d_{i-1}, d_i$ are positive integers.
\end{definition}

Layers in neural networks often have more structure than just a matrix-vector multiplication as above (e.g., convolutional layers). However, they may admit a description as a special form of a fully connected layer. We call a network \emph{fully connected} if all its layers are fully connected. The first $r$ layers are the \emph{hidden layers}, and layer $r+1$ is the \emph{output layer}. 

\begin{definition}
A \emph{neuron} is a function determined by the corresponding weight matrix and activation function. Particularly, the $j$-th neuron of layer $i$ is the function $\eta$ given by
\[ \eta(x) = \sigma(A_{j}^{(i)}x + b_j^{(i)}), \]
where $A_{j}^{(i)}$ and $b_j^{(i)}$ denote, respectively, the $j$-th row of $A^{(i)}$ and $j$-th coordinate of $b^{(i)}$.
\end{definition}

\begin{definition}
Let $\ell = d_{i-1}$ and $A_{j}^{(i)}$ be described as $(a_1, a_2, \ldots, a_\ell)$. The \emph{signature} of the $j$-th neuron in layer $i$ is the tuple
\begin{equation}\label{eq:sig}
\left(\frac{a_1}{a_1} = 1, \frac{a_2}{a_1},\ldots ,\frac{a_\ell}{a_1}  \right).
\end{equation}
\end{definition}

\begin{definition}
Let $\mathcal{V}(\eta;x)$ denote the value that neuron $\eta$ takes with $x \in \mathcal{X}$ before applying $\sigma$. If $\mathcal{V}(\eta;x) > 0$ then $\eta$ is \emph{active}. When $\mathcal{V}(\eta;x) = 0$, $\eta$ is \emph{critical}, and we call $x$ a \emph{critical point} for $\eta$\footnote{Carlini et al.'s ``critical point'' is ``being critical'' in our definition. 
Carlini et al.'s ``witness for a neuron being at a critical point'' is ``a critical point'' in our definition.}. Otherwise, it is \emph{inactive}. The state of $\eta$ on input $x$ (i.e., active, inactive, or critical) is denoted by $\mathcal{S}(\eta;x)$.
\end{definition}

\begin{definition}
Let $x \in \mathcal{X}$. The \emph{linear neighbourhood} of $x$ is the set
\[ \{ u \in \mathcal{X} \mid \mathcal{S}(\eta;x) = \mathcal{S}(\eta;u) \text{\ for all neurons $\eta$ in the network\,} \}. \]
\end{definition}

\begin{definition}
The \emph{architecture} of a fully connected neural network is described by specifying its number of layers along with the dimension $d_i$ (i.e., number of neurons) of each layer $i =1,\cdots, r+1$. We say that $d_0$ is the dimension of the inputs to the neural network, whereas $d_{r+1}$ gives the number of outputs of the network. A neural network has $N = \sum_{i=1}^{r} d_{i}$ neurons.
\end{definition}

As in~\cite{carlini2020cryptanalytic}, we specify the architecture of a neural network by enumerating the dimensions of its layers. For example, the eight-hidden-layer network for classifying the CIFAR10 dataset showcased in this paper has the architecture
\[ 3072-256^{(8)}-10. \]

Let $F_{i}$ denote the function that computes the first $i$ layers of the DNN after the ReLUs, i.e., $F_{i} = \sigma \circ f_{i} \circ \cdots \circ \sigma \circ f_{1}$. By definition, all neurons remain in the same state when evaluating the DNN with an input in the linear neighborhood of $x\in \mathcal{X}$. Following the explanation in \cite{carlini2020cryptanalytic}, for any such point $x'$, we have that
\begin{align*}
    F_{i}(x') &= I^{(i)}(A^{(i)} \cdots (I^{(2)}(A^{(2)}(I^{(1)}(A^{(1)}x' + b^{(1)})) + b^{(2)}) \cdots + b^{(i)}) \\
                &= I^{(i)}A^{(i)} \cdots I^{(2)}A^{(2)}I^{(1)}A^{(1)}x' + \beta \\
                &= \Gamma x' + \beta,
\end{align*}
where $I^{(j)}$ are $0-1$ diagonal matrices with a $0$ on the diagonal's $k$-th entry when neuron $k$ at layer $j$ is inactive and $1$ on the diagonal's $k$-th entry when that neuron is active. That is, in the linear neighborhood of an input $x$, we can ``collapse'' the action of various contiguous layers into an affine transformation. If we make a change $\Delta$ to the input, we can observe the corresponding change in the value of the neurons:
\[ F_{i}(x + \Delta) - F_{i}(x) = \Gamma(x + \Delta) + \beta - (\Gamma(x) + \beta) = \Gamma\Delta. \]
This $\Delta$ must be such that $x + \Delta$ is in the linear neighborhood of $x$.

\begin{figure}[!ht]
    \centering
    \includegraphics[width=8cm]{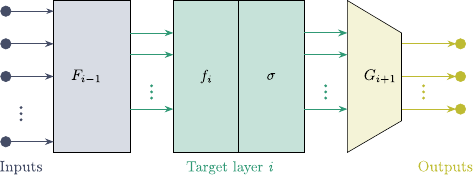}
    \caption{Representation of the DNN according to the recovered part $F_{i-1}$, current target layer $i$ and unknown part $G_{i+1}$.}
    \label{fig:Network}
\end{figure}

Assume that we fully know the first $i-1$ layers, and we are currently recovering layer $i$. Let $F_{i-1}$ and $G_{i+1}$ represent, respectively, the fully recovered and non-recovered part of the DNN, i.e.,
\[ f = \underbrace{f_{r+1} \circ \sigma \circ \cdots \circ \sigma \circ f_{i+1}}_{G_{i+1}} \circ \sigma \circ f_i \circ \underbrace{\sigma \circ f_{i-1} \circ \cdots \circ \sigma \circ f_{1}}_{F_{i-1}}. \]
Then, the neural network can be depicted as in \autoref{fig:Network}. Furthermore, if we restrict inputs $x'$ to be in the linear neighbourhood of $x$, we can collapse $F_{i-1}$ and $G_{i+1}$ as
\[ F_{i-1}(x') = F^{(i-1)}_{x}x' + b_{x}^{(i-1)} \quad \text{and} \quad  G_{i+1}(x') = G^{(i+1)}_{x}x' + b_{x}^{(i+1)}, \]
respectively. We use the subscript in the collapsed matrices and bias vectors to indicate that they are defined in the linear neighborhood of $x$.

\begin{definition}
We call the $j$-th row of  $G^{(i+1)}_x$ the \emph{output coefficients} for the $j$-th output of the DNN.
\end{definition}

\subsection{Problem Statement and Assumptions}
\label{sec:prob}

The parameters $\theta$ of a DNN $f_{\theta}$ are the concrete assignments to the weights and biases. Following the setting in \cite{carlini2020cryptanalytic}, in a \emph{model parameter extraction attack}, an attacker generates queries $x$ to an oracle $\mathcal{O}$ which returns $f_{\theta}(x)$. The goal of the attacker is to obtain a set of parameters $\hat{\theta}$ such that $f_{\hat{\theta}}(x)$ is as similar as possible to $f_{\theta}(x)$.

We focus on the attack presented in  \cite{carlini2020cryptanalytic}. As mentioned in \autoref{sec:Introduction_OverviewOfAttack}, Carlini et al.'s attack recovers the parameters layer by layer in two steps. The first one finds multiples of the parameters, particularly the signatures of the neurons as defined in~\autoref{eq:sig}. Since the signature consists of ratios of pairs of weights, negating all these weights simultaneously will preserve the signature but nonlinearly change the outputs of this neuron's ReLU. Consequently, before we can peel off a layer of neurons, we have to determine for each one of its neurons separately whether the weights and biases are one possible vector of values or its negated vector. That is, we must get a sign of that neuron's weight. The second step recovers these signs for all neurons in the current layer. In this paper, we specifically focus our attention on the latter half, which was the exponential-time bottleneck in~\cite{carlini2020cryptanalytic}, under the assumption that the signatures are already known.

The following are the assumptions we have regarding the oracle and the capabilities of the attacker:
\begin{itemize}
    \item \textbf{Full-domain inputs}. We can feed arbitrary inputs from $\mathcal{X} = \mathbb{R}^{d_0}$.
    \item \textbf{Complete outputs}. We receive outputs directly from $f$ without further processing.
    \item \textbf{Fully connected network and ReLU activations}. The network is fully connected, and all activation functions are the ReLU function.
    \item \textbf{Fully precise computations}.\label{item:assu_infinite_precision} All computations are done with infinite-precision arithmetic.
    \item \textbf{Signature availability and uniqueness}. We have access to the signature of each neuron. Also, we assume that no two signatures are the same.
\end{itemize}
All but the last one are also assumptions in \cite{carlini2020cryptanalytic}. Regarding full precision, we remark that computations by both the oracle and the attacker enjoy this characteristic. Carlini et al.\ assume single-output DNNs. However, we allow for multiple outputs, which enhances the performance of our techniques and is incidentally also more realistic. Finally, if a full attack following \cite{carlini2020cryptanalytic} using our sign recovery methods is to be performed, we must also assume \emph{knowledge of the architecture}. This assumption is required to apply the methods in \cite{carlini2020cryptanalytic}.

\subsection{Carlini et al.'s Differential Attack}

We now present a high-level description of the techniques by Carlini et al.~\cite{carlini2020cryptanalytic}.

\subsubsection{Finding critical points}

\begin{figure}[!h]
\centering
    \begin{subfigure}[b]{0.50\textwidth}
		\centering
        \includegraphics[width=\linewidth]{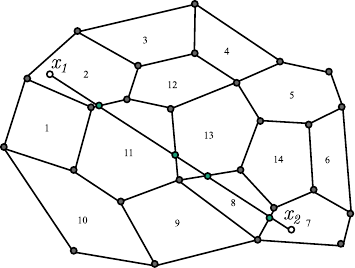}
	\end{subfigure}
    \hfill
	\begin{subfigure}[b]{0.47\textwidth}
		\centering
        \resizebox{5cm}{!}{\begin{tikzpicture}[line width=0.5mm,node distance=0mm]

\draw (0.0,10) -- (0.0,0.0) -- (10,0.0);
\node (y_lbl) at (0.0,5.0) {};
\node (x_lbl) at (5.0,0.0) {};
\node [left=of y_lbl,anchor=east]   {\LARGE $f^*(\lambda)$};
\node [below=of x_lbl,anchor=north] {\LARGE $\lambda$};

\draw (0.5,7.5) -- (1.5,8.0) -- (3.5,4.8) -- (5.5,4.5) -- (8.5,6.5) -- (9.5,5);

\draw [dashed] (0.5,0) -- (0.5,10);
\draw [dashed] (1.5,0) -- (1.5,10);
\draw [dashed] (3.5,0) -- (3.5,10);
\draw [dashed] (5.5,0) -- (5.5,10);
\draw [dashed] (8.5,0) -- (8.5,10);
\draw [dashed] (9.5,0) -- (9.5,10);

\node at (1.0,9.0)  {\LARGE \rotatebox{90}{neigh.\  2}};
\node at (2.5,9.0)  {\LARGE \rotatebox{90}{neigh.\ 11}};
\node at (4.5,9.0)  {\LARGE \rotatebox{90}{neigh.\ 13}};
\node at (7.0,9.0)  {\LARGE \rotatebox{90}{neigh.\  8}};
\node at (9.0,9.0)  {\LARGE \rotatebox{90}{neigh.\  7}};
\node at (0.5,10.5) {\LARGE $x_1$};
\node at (9.5,10.5) {\LARGE $x_2$};

\end{tikzpicture}}
	\end{subfigure}
\caption{The input space can be partitioned into linear neighborhoods, and the output function displays abrupt changes in behavior when moving across their borders.}
\label{fig:input_space}
\end{figure}
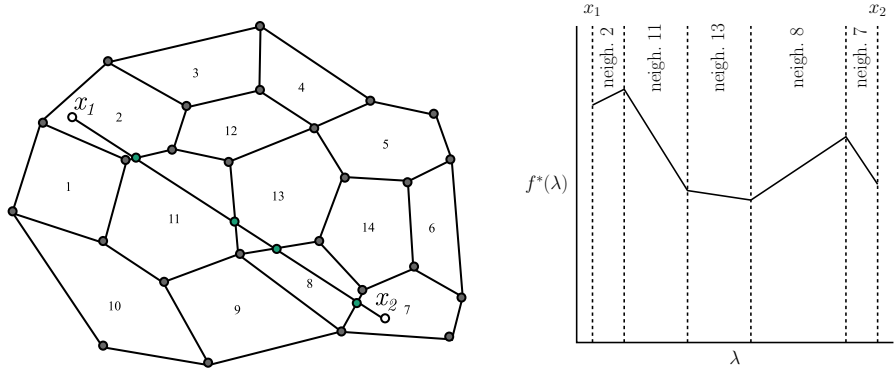

To discover critical points, Carlini et al.\ analyze the function induced by a DNN when an input $x_1$ is linearly transformed into another input $x_2$. For the sake of simplicity and without loss of generality, we will assume in the following that the network has a single output. 
Let $x_1, x_2 \in \mathbb{R}^{d_0}$ and $\mu:[0, 1] \rightarrow \mathbb{R}^{d_0}$ defined as
\[ \mu:\lambda \mapsto x_1 + \lambda (x_2 - x_1) \]
be the linear transformation of $x_1$ into $x_2$. This induces an output function on the DNN,
\[ f^*(\lambda) := f( \mu(\lambda) ) ,\]
which is a piecewise linear function with first-order discontinuities precisely when one of the neurons is toggling between active/inactive states. As shown in \autoref{fig:input_space}, we can identify the first-order discontinuities and revert the mapping $\mu$ to recover the point at which the line from $x_1$ to $x_2$ intersects a boundary between linear neighborhoods, which is a critical point for some neuron. 

In practice, it suffices to measure the slope of the graph in \autoref{fig:input_space} at different points and extrapolate to find the critical point, all while checking that there are no other abrupt changes in behavior in-between. We refer to \autoref{sec:findCriticalPoints} for the fully-detailed algorithm.

At this point, we do not yet know to which layer each critical point belongs, but by sampling enough pairs $x_1,x_2$ we expect to eventually find multiple critical points for every neuron in the network.

\subsubsection{Finding signatures}

The input to the DNN is also the input to the first hidden layer, and we have full control over it. Let $\eta$ be a neuron with weights $(a_1,\dots,a_{\ell})$. Given $x^* \in \mathbb{R}^{d_0}$ a critical point for $\eta$, we can query $ \alpha_{i,-} = \frac{\partial f}{\partial e_i} (x^* - \varepsilon e_i)$ and $\alpha_{i,+} = \frac{\partial f}{\partial e_i} (x^* + \varepsilon e_i)$, where $\{e_i\}_{i=1}^{{d_0}}$ is the canonical basis of $\mathbb{R}^{d_0}$ and $\varepsilon$ is a small real number. Since $x^*$ is a critical point, only either $\alpha_{i,-}$ or $\alpha_{i,+}$ will have $\eta$ in its active state assuming that $\varepsilon$ is sufficiently small so that no other neuron toggles. The difference $\alpha_{i,+} - \alpha_{i,-}$ contains the gradient information moving from the input coordinate $i$ through neuron $\eta$ and to the output; the gradient information through all other neurons cancel out. In other words, this difference is a multiple of $a_i$ given by the other layers of the DNN. Dividing out by another coordinate eliminates the multiplicative factor, i.e., $(\alpha_{i,+} - \alpha_{i,-}) / (\alpha_{k,+} - \alpha_{k,-}) = a_i/a_k$. If we fix $k=1$, the signature \eqref{eq:sig} is recovered. We denote by $\hat{A}_j^{(i)}$ the signature of the $j$-th neuron in layer $i$ and $\hat{A^{(i)}}$ the matrix whose $j$-th row is $\hat{A}_j^{(i)}$. Critical points for the same neuron in the target layer $1$ will yield the same signature, while critical points for neurons in other layers will generate different signatures. This allows us to decide which signatures correspond to a layer-$1$ neuron (i.e., those signatures appearing with repetitions). See \cite{carlini2020cryptanalytic} for details on this.

After peeling off layers, we can also determine if a signature corresponds to a neuron in the current target layer by observing repetitions. However, starting from layer $2$ we no longer have full control of the layer's input, and applying the method above is not possible (we cannot change one coordinate at a time). To overcome this, in layer $i > 1$, we sample $d_i + 1$ directions $\delta_k \sim \mathcal{N}(0, \varepsilon I_{d_0}) \in \mathbb{R}^{d_0}$, and let $\{ y_k \} = \{ \partial^2 f(x^*) / \partial \delta_1 \partial \delta_k \}_{k=1}^{d_i}$ and $h_k = F_{i-1}(x^* + \delta_k)$. The signature is then given by the vector $a$ such that $\langle h_k, a \rangle = y_k$. This, however, yields partial signatures (since the ReLUs in the previous layer set negative values to zero). Different critical points for the same neuron yield different partial signatures. 
Each partial signature will yield a different set of coordinates and with enough partial signatures, we can reconstruct the full one.

\subsubsection{Finding signs using the freezing method}

We now consider the problem of finding the signs of a given layer. If the network is not expanding, this problem is easy for the first hidden layer. Indeed, for target neuron $k$ in layer $i$, we can find a wiggle $\Delta_k$ in the input space that produces a wiggle $\pm e_k$ in the first hidden layer, where $e_k\in\mathbb R^{d_i}$ is a basis vector in the $k$-th direction, by solving a system of $d_1$ equations in $d_0$ variables given by the first layer's weight matrix. For any $x$ in the input space, the outputs of the neural network at $x$ and $x+\Delta_k$ will be equal if the $k$-th neuron is inactive in the linear neighborhood of $x$ (since the $k$-th neuron is suppressed by the ReLU whereas all other neurons remain unchanged by construction), and will be different otherwise. We refer to this simple sign-recovery technique as \Freeze.

There are some scenarios where the same method can be applied to deeper layers. Note that in the linear neighborhood of any input $x$, the mapping from the $d_0$-dimensional input space to the space of values entering layer $i$ is an affine mapping since there are no ReLUs that flip from active to inactive or vice versa. The rank of this mapping determines in how many linearly independent directions we can slightly perturb the inputs to layer $i$ when we consider arbitrary perturbations of the input $x$. If the rank is high enough, we can still expect to find preimages for each of the basis vectors and can apply the \Freeze method.

 In general, however, our ability to change the inputs to a deep hidden layer is severely limited, since about half of the neurons in each layer are expected to be suppressed, and thus, the rank of the affine mapping decreases as we move deeper into the network. This issue is pointed out in~\cite{carlini2020cryptanalytic}, where it is claimed that the \Freeze method can only be applied in networks that are ``\emph{sufficiently contracting}" (that is, layer size should decrease by roughly a factor of 2 in each layer in order to compensate for the rank loss). In networks that are not sufficiently contractive, Carlini et al.\ use instead the much more expensive technique of exhaustive search over all the possible sign combinations (as described in~\cite[Section 4.4]{carlini2020cryptanalytic}) that is exponential in time, which makes their attack feasible only for toy examples of non-contracting networks.

\section{Our New Sign-Recovery Techniques}
\label{sec:sign_recovery_techniques}

As just mentioned, the diminishing control over the inputs to a layer is a significant problem that makes sign recovery in deeper layers harder. To better understand our proposed solutions, we first describe this problem in greater detail.

Let $x \in \mathbb{R}^{d_0}$ be an input to the DNN. Recall that $F^{(i-1)}_{x}$ and $G^{(i+1)}_{x}$ are the collapsed matrices for $x$ corresponding to the already recovered and unknown part of the DNN, respectively.

\begin{definition}\label{def:space-of-control}
The \emph{space of control for layer $i$ around input $x$}, denoted by $V_x^{(i-1)}$, is the range of the linear transformation $F_x^{(i-1)}$. The dimension of this space is called the \emph{number of degrees of freedom for layer $i$ with input $x$} and is denoted by $d_x^{(i-1)}$.
\end{definition}

The space of control is the vector space containing all possible small changes at the input to layer $i$ and, by the definition, $d_x^{(i-1)} = \textrm{rank}(F_x^{(i-1)})$. Due to the ReLUs in layers $1$ to $i-1$ making neurons inactive, for a fixed $x$, the number of degrees of freedom remains equal or decreases with increasing $i$. To see this, consider an input $x$ making half of the $d$ neurons in layer 1 be active. The matrix $F_x^{(1)}$ would then have rank $d/2$, and particularly, the rows corresponding to inactive neurons are the zero vector. If layer two has $d'<d/2$ active neurons (with the same input $x$), the matrix $F_x^{(2)}$ has rank $d'$, and therefore the number of degrees of freedom is also $d'$. Now, if $d' \geq d/2$, $F_x^{(2)}$ has rank $d/2$ (since the rank of $F_x^{(1)}$ cannot be increased when multiplied by another matrix). We have the same situation for the subsequent layers: the rank can never be increased once it has decreased. In fact, the number of degrees of freedom at layer $i$ is typically determined by the minimum number of active neurons per layer among all layers $1$ to $i-1$, but in some cases, it can be strictly smaller.

\begin{figure}[!h] 
\begin{center}
	\begin{subfigure}[b]{25mm}
		\centering
        \includegraphics[width=\linewidth]{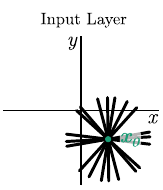}
        \caption{}
        \label{fig:rank-example-input}
	\end{subfigure}
\hfill
	\begin{subfigure}[b]{25mm}
		\centering
        \includegraphics[width=\linewidth]{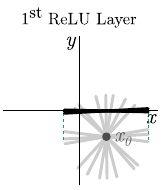}
        \caption{}
        \label{fig:rank-example-relu01}
	\end{subfigure}
\hfill
	\begin{subfigure}[b]{25mm}
		\centering
        \includegraphics[width=\linewidth]{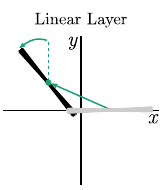}
        \caption{}
        \label{fig:rank-example-linear02}
	\end{subfigure}
\hfill
	\begin{subfigure}[b]{25mm}
		\centering
        \includegraphics[width=\linewidth]{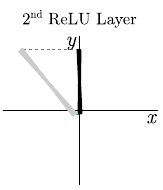}
        \caption{}
        \label{fig:rank-example-relu02}
\end{subfigure}
\end{center}
\caption{Intuition for the space of control (\autoref{def:space-of-control}). Consider a network with 2-dimensional input and two hidden layers with two neurons in each. \subref{fig:rank-example-input}~For the first hidden layer, the space of control is the full 2-dimensional space. We can move around the input $x_0$ in any direction. \subref{fig:rank-example-relu01}~After the first hidden layer, if one ReLU is positive and the other negative, we lose one dimension of control. 
\subref{fig:rank-example-linear02}~The linear transformation in the second layer will rotate, translate, and scale the space of control. 
\subref{fig:rank-example-relu02}~Therefore, after the ReLU's in the second layer (if again one is positive and one negative), we are still left with a one-dimensional space of control. Note that with no rotation, the space of control collapses into a point after these two hidden layers.}
\label{fig:rank-example}
\end{figure}

Consider a DNN with input dimension $d$ and the same width $d$ in all its hidden layers. Assume further that each neuron has probability $1/2$ of being active. Initially, the number of degrees of freedom is equal to the dimension $d$ of the input to the DNN. Then, the first hidden layer will have, on average, half of its neurons active, which drops the number of degrees of freedom at the input of layer 2 to $d/2$. We can think of the ReLU's projecting the space of control onto a space determined by the active neurons. One may think that half of those $d/2$ degrees of freedom will again be lost in the second layer due to the fact that half of its ReLU's will be inactive, ending with $d/4$ degrees of freedom. However, before going to the ReLUs on that layer, the space of control for layer 2 is typically rotated by $A^{(2)}$. This rotation may make many coordinates survive the projection of the ReLUs in the second layer. We may still lose some degrees of freedom, but not as many as half of them in each successive layer. Due to this effect of the linear transformation, the number of degrees of freedom will typically stabilize after the first few layers. \autoref{fig:rank-example} depicts this phenomenon in a two-dimensional space, and \autoref{table:positive-relus} shows the average number of active neurons and the average number of degrees of freedom in the 8 successive hidden layers of the actual CIFAR10 network we attack.

We now describe our methods for sign recovery with the loss of degrees of freedom in mind. Recall that in the context of the attack, when recovering the signs for layer $i$, we fully know $F_{i-1}$, we know $f_i$ up to a sign per neuron, and $G_{i+1}$ is completely unknown. To simplify our notation, we may drop the subscript $x$ from the collapsed matrices, space of control, and number of degrees of freedom if $x$ is clear by the context.

\subsection{SOE Sign-Recovery}\label{sec:soe}

We first describe a method for sign recovery in cases where the number of degrees of freedom is sufficiently large. We refer to the method as SOE since it relies on solving a System Of Equations. This method is superficially similar to \Freeze, but uses different equations and a different set of variables (in the case of~\cite{carlini2020cryptanalytic}, the variables referred to the direction we have to follow in input space to freeze all the neurons except the targeted one, while in our SOE technique the variables refer to the coefficients of the output function $G_{i+1}$ at some randomly selected point $x$). Our technique is more efficient in both its query and time complexities since Carlini et al.\ had to solve a different system of equations for each targeted neuron, while in our SOE technique, one system of linear equations can simultaneously provide the signs of all the neurons in the current layer.


We assume without loss of generality that the network has a single output (additional outputs can be simply ignored).

As before, let $I^{(i)}_x$ be the matrix representing the ReLU at layer $i$ on input $x$. The equation for the change in output under an arbitrary sequence of changes in input $\Delta_k$, can then be written as
\[  f(x + \Delta_k) - f(x) = G_x^{(i+1)}I^{(i)}_x A^{(i)}F^{(i-1)}_x \Delta_k . \]

Let $y_k = A^{(i)}F^{(i-1)}_x \Delta_k$ and $c = G_x^{(i+1)}I^{(i)}_x$. If the left-hand side is observed to take values $z_k$ through direct queries, the equation can be rewritten as
\[ c\cdot y_k = z_k, \]
which can be regarded as a system of equations where $c$ is a vector of variables. Because of the ReLU, if neuron $j$ is inactive on input $x$ we will necessarily have $c_j = 0$, so after obtaining $d_i$ equations, we can solve the system and determine which neurons are inactive around $x$ and hence the appropriate choice of signs for the current layer.

\begin{remark}
In the context of the attack, we can only recover $y_k$ up to a global scaling of each entry (including possibly a sign flip), but this results in a system of equations where
the solution has the same set of variables vanishing.
\end{remark}

\subsubsection{Oracle Calls/Time}

This method is optimal in terms of queries since it only requires $d_i+1$ queries to solve a layer of size $d_i$ (namely, it must query $f(x)$ and $f(x+\Delta_k)$ for $d_i$ values of $k$). Once the queries have been performed, the time complexity of the attack comes from solving a system of equations, which can be done in $\mathcal O(d_i^3)$ with standard methods.


\subsubsection{Limitations}

Each choice of $\Delta_k$ produces a new equation, so we attempt to gather more equations until the system is uniquely solvable. However, the equations that are obtained may not all be linearly independent. In fact, since the $y_k$ all lay in $A^{(i)}(V_x^{(i-1)})$, we will only obtain enough linearly independent equations if $d^{(i-1)}_x \geq d_i$. This is likely to be the case if the layer size is steadily contracting by a factor of $2$, but there is an important exception for the first two hidden layers. Indeed, as long as the network is not expanding, the linear map corresponding to the first hidden layer can usually be inverted. This makes it easy to find an $x$ at which all first-layer neurons are active and hence $d^{(1)}_x = d_1 \geq d_2$. Therefore, the method applies straightforwardly for the first two hidden layers on the sole condition of non-expansiveness and is likely to succeed in subsequent layers only if they each contract by a factor close to $2$.


\subsection{\NeuronWiggle Sign-Recovery}
\label{sec:neuron-wiggle}


The method presented here does not have the strict constraints on the network's architecture as the one above, since its performance gradually degrades as the number of degrees of freedom diminishes, whereas in SOE the system of equations abruptly changes from solvable to unsolvable.

\begin{definition}
    A \emph{wiggle at layer $i$} is a vector $\delta \in \mathbb{R}^{d_{i-1}}$ of differences in the value of the neurons in that layer.
\end{definition}

Recall that the recovered weight matrix for layer $i$ is $\hat{A}^{(i)} \in \mathbb{R}^{d_i \times d_{i-1}}$ and let its $k$-th row be $\hat{A}_{k}^{(i)}$. For a wiggle $\delta$ at layer $i$, neuron $k \in \{1,\dots,d_i\}$ in that layer changes its value by $e_k = \langle \hat{A}_{k}^{(i)}, \delta \rangle$. Consider one output of the DNN and let $(c_1,\dots,c_{d_i})$ be its corresponding vector of output coefficients (i.e., its corresponding row vector in $G^{(i+1)}$). If we ``push'' the differences $e_k$ through the remaining layers, the difference in the output is
\begin{equation}
\label{eq:wiggling_outputDifferenceDNN}
    \sum_{k \in I} c_ke_k,
\end{equation}
where $I$ contains the indices of all active neurons at layer $i$. We want to recover the sign of neuron $j$. If this neuron is active, the output difference contains the contribution of $e_j$. If the neuron is inactive, the ReLU ``blocks'' $e_j$ and it does not contribute. When $c_je_j$ is sufficiently large, we can detect whether it is present in \eqref{eq:wiggling_outputDifferenceDNN} and use this information to recover the sign of the neuron. This is best achieved when $\delta$ is a wiggle that maximizes the change in value for that neuron, i.e., $\|\hat{A}^{(i)}\delta\|_{\infty} = |e_j|$. The crucial property we use here is that maximizing the \emph{size} of the wiggle produced by a linear expression does not require knowledge of its sign - if we negate the expression we get a wiggle of the same size but in the opposite direction. We now show how to compute such a maximal wiggle and how to recover the target neuron's sign.

\subsubsection{Compute Target Neuron Wiggle}

Let $\delta \in \mathbb{R}^{d_{i-1}}$ be parallel to $\hat{A}_{j}^{(i)}$ (i.e., all coordinates of $\delta$ have either the same or opposite sign to the corresponding coordinate of $\hat{A}_{j}^{(i)}$). Then, all summands in the dot product $\langle \hat{A}_{j}^{(i)}, \delta \rangle$ have the same sign. If no other row of $\hat{A}^{(i)}$ is a multiple of $\hat{A}_{j}^{(i)}$, with very high probability $|\langle \hat{A}_{j}^{(i)}, \delta \rangle| > |\langle \hat{A}_{k}^{(i)}, \delta \rangle|$, for all $k \neq j$. That means $\|\hat{A}^{(i)}\delta\|_{\infty} = |\langle \hat{A}_{j}^{(i)}, \delta \rangle|$. Hence, the change of value for neuron $j$ can be maximized if the wiggle is parallel to $\hat{A}_{j}^{(i)}$.

\begin{figure}[!ht]
    \centering
    \resizebox{4.5cm}{!}{\begin{tikzpicture}[line width=0.5mm,node distance=0mm]

\draw (-3,-2) -- (3,-2) -- (6,2) -- (0,2) -- (-3,-2);

\draw[black,fill=black] (0,0) circle (0.5mm);
\draw [->]              (0,0) -- (4,3);

\draw [dashed,->] (0,0) -- (4,0);
\draw [dashed,line width=0.2mm]    (4,3) -- (4,0);

\draw [dashed] (0,0) ellipse (1 and 0.65);
\draw [dashed] (0,0) -- (-0.75,-0.45);

\draw [->](0,0) -- (1,0);

\node (sc_aux) at (0,-2) {};
\node [below=of sc_aux] {\Large Space of control};
\node (sig_aux) at (4,3) {};
\node [right=of sig_aux] {\Large $\hat{A}_j^{(i)}$};
\node [anchor=north west] at (1,0) {\Large $\delta$};
\node [] at (-0.5,0) {\Large $\varepsilon$};

\end{tikzpicture}}
    \caption{Computing a wiggle that maximizes the change in the target neuron.}
    \label{fig:wiggle_projectionOntoSpaceOfControl}
\end{figure}
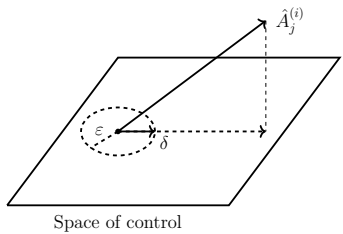

Recall that $V^{(i-1)}$ is the space of control for layer $i$ given an input $x$. We project $\hat{A}_{j}^{(i)}$ onto $V^{(i-1)}$ and get $\delta$ by scaling this projection to have a sufficiently small norm $\varepsilon^{(i-1)}$; see \autoref{fig:wiggle_projectionOntoSpaceOfControl}. Finally, we get the input difference $\Delta \in \mathbb{R}^{d_0}$ that generates $\delta$ by finding a pre-image of $\delta$ under $F^{(i-1)}$.

\subsubsection{Recover Target Neuron Sign}

We want to recover the sign for $\eta_j$, the $j$-th neuron in layer $i$. Let $x^*$ be a critical point for $\eta_j$ and let $\Delta \in \mathbb{R}^{d_0}$ generate the wiggle $\delta$ (at layer $i$) that maximizes the change in value for that neuron. Assume the sign of $\eta_j$ to be positive. Then, the signs of the recovered weights $\hat{A}_{j}^{(i)}$ are the same as those of the real weights $A_{j}^{(i)}$. This implies that also the coordinates of $\delta$ have the same sign as those of $A_{j}^{(i)}$ and $e_k = \langle A_{k}^{(i)}, \delta \rangle$ has a positive value. Assume the DNN has a single output. Since $x^*$ is a critical point for $\eta_j$, evaluating the DNN at $x^* + \Delta$ makes $\eta_j$ active, thus
\[ f(x^* + \Delta) - f(x^*) = c_je_j + \sum_{k \in I \setminus \{j\}} c_ke_k, \]
where $I$ contains the indices of all active neurons at layer $i$. It is necessary that $\Delta$ changes the state of neuron $\eta_j$ only. Evaluating at $x^* - \Delta$ makes the wiggle $\delta$ have opposite signs to those in $A_{j}^{(i)}$. Then, all differences $e_k$ also have opposite signs (compared to evaluating at $x^* + \Delta$). In this case, $\eta_j$ becomes inactive and we have that
\[ f(x^* - \Delta) - f(x^*) = -\sum_{k \in I \setminus \{j\}} c_ke_k. \]
Now assume that $\eta_j$ has a negative sign. Then, the wiggle $\delta$ will have opposite signs to those in $A_{j}^{(i)}$ and following a similar analysis as above, we get that 
\[ f(x^* + \Delta) - f(x^*) = -\sum_{k \in I \setminus \{j\}} c_ke_k \]
and
\[ f(x^* - \Delta) - f(x^*) = c_je_j + \sum_{k \in I \setminus \{j\}} c_ke_k. \]
So, in order to find the sign we need to distinguish whether $c_je_j$ contributes to the output difference with $x^* + \Delta$ or $x^* - \Delta$.

Let $L = f(x^* - \Delta) - f(x^*)$ and $R = f(x^* + \Delta) - f(x^*)$ denote, respectively, the output difference to the left and right of $x^*$. We decide $c_je_j$ appears on the left, i.e., the sign of the neuron is $-1$, if $|L| > |R|$. Otherwise, we decide the sign to be $+1$. Since $c_je_j \neq 0$, it is not possible that $|L| = |R|$.

If $c_je_j$ and $\sum_{k} c_ke_k$ have the same sign, then $\left| c_je_j + \sum_{k} c_ke_k \right| > \left| -\sum_{k} c_ke_k \right|$ always holds and the decision on the sign is also always correct. If $c_je_j$ and $\sum_{k} c_ke_k$ have opposite signs, however, an incorrect decision may occur. This wrong decision happens when $\left| -\sum_{k} c_ke_k \right| > \left| c_je_j + \sum_{k} c_ke_k \right|$. Then, it is necessary that $|c_je_j| > 2|\sum_{k} c_ke_k|$ to make a correct decision.

Recall that a given input to the DNN defines a particular matrix $G^{(i+1)}$. That is, different inputs define different coefficients for the output of the DNN. We refer to this fact as \emph{output randomization} and exploit it to overcome the problem of making a wrong sign decision. We find $s$ different critical points for neuron $\eta_j$; each point defines different output coefficients. We expect that the majority of these points define coefficients such that $c_je_j$ and $\sum_{k} c_ke_k$ fulfill the conditions for making a correct decision. For each critical point, we compute the wiggle $\delta$, its corresponding input difference $\Delta$, and make the choice for the sign. Let $s_{-}$ and $s_{+}$ denote, respectively, the number of critical points for which the sign is chosen to be $-1$ and $+1$. Also, let the \emph{confidence level} $\alpha$ for $-1$ be $s_{-}/s$ and $s_{+}/s$ for $+1$. Then, decide the sign to be $-1$ if $s_{-} > s_{+}$ and its confidence level is greater than a threshold $\alpha_0$. If $s_{+} > s_{-}$ and its confidence level is greater than $\alpha_0$, decide $+1$. Otherwise, no decision on the sign is made. When the latter happens, we may try to recover the sign with additional critical points. 

In our experiments in \autoref{sec:experiments}, very few wrong signs were initially produced by testing 200 critical points for each neuron in the CIFAR10 network; all of them were known to be problematic due to their low confidence level, and they are all fixable by testing more critical points.
Note that as the number of neurons in the network increases, we expect the neuron wiggling technique to get even better since in higher dimensions vectors tend to be more orthogonal to each other, and thus the ratio between the sizes of the wiggles in the targeted neuron and in other neurons should increase.

So far, we assumed a single output for the DNN. When it has multiple outputs, we can use the Euclidean norm over the vector of outputs to compare $L$ and $R$, which is beneficial for this method. This is because each critical point randomizes the coefficients of multiple outputs, and the probability of multiple outputs having simultaneously ``bad'' coefficients (which may lead to wrong decisions) is lower.

\subsubsection{Oracle Calls/Time}

Recall that layer $i$ has $d_i$ neurons. To recover the sign of a single neuron, for $s$ critical points, we compute $\Delta$ and compare $L$ with $R$. Computing $\Delta$ requires no oracle queries: it only requires linear algebra operations to find a critical point $x^*$, project $\hat{A}_j^{(i)}$ onto $V^{(i-1)}$ and find $\Delta$. Particularly, matrix multiplications and matrix inversions. The size of the matrices involved is given by the number of inputs $d_0$ and the number of neurons $d_i$. So, the time complexity is $\mathcal O(d^3)$ operations, where $d = \max(d_0, d_i)$. Comparing $L$ with $R$ requires querying the oracle on $x^*$, $x^* - \Delta$ and $x^* + \Delta$. Computing $L = \|f(x^* - \Delta) - f(x^*)\|$ and $R = \|f(x^* + \Delta) - f(x^*)\|$ requires $\mathcal O(d_{r+1})$ operations, where $d_{r+1}$ is the number of outputs of the DNN. Thus, we require $3s$ queries and $\mathcal O(sd^3)$ operations for a single neuron. In total, we require $3sd_i$ queries and $\mathcal O(sd_id^3)$ operations to recover the sign of all neurons in layer $i$.

Appendix~\ref{sec:neuronWiggleUnitaryBalancedNetworks} contains a back-of-the-envelope estimation of the signal-to-noise ratio for each critical point we test, showing that even in expansive networks we only require a relatively small number of experiments $s$ to make a good guess.

\subsubsection{Limitations}
We can think of $c_je_j$ as a signal and $\sum_{k} c_ke_k$ as noise. We have seen that when the signs of the signal and the noise are different, we may wrongly decide the sign of a neuron. This happens when the signal is not big enough compared to the noise. Particularly, if the number of neurons in layer $i$ is too large compared to the degrees of freedom (for a particular input $x$), the signal may be really weak with respect to the noise. That means this technique may not work with DNNs which have at least one hidden layer with a large expansion factor compared to the smallest hidden layer or the number of inputs. However, we had no trouble recovering the signs when several successive hidden layers had twice the number of neurons compared to the dimension of the input space.

Also, this method may not be suitable for DNNs with a small number of neurons in the target layer. The probability of two random vectors being perpendicular decreases in low-dimensional spaces. Therefore, the wiggle for the target neuron may produce a sensibly large change for other neurons as well (those with weight vectors somewhat parallel to that of the target neuron). In this situation, the contribution of the other neurons may counteract that of the target neuron.

Finally, this method leverages \emph{output randomization}. There is no output randomization when recovering the last hidden layer.  
If there are neurons in that layer with bad (constant) output coefficients, their corresponding sign will always be incorrectly recovered. The method in the next section exploits this lack of output randomization to recover the signs in the last hidden layer.

\subsection{\LastLayer Sign-Recovery}
\label{sec:last-layer}


The method presented here recovers the sign of neurons in the last hidden layer. The output of the DNN is produced by the affine transformation $f_{r+1}$ \emph{without} subsequent ReLUs. Therefore, in layer $r$ (the last hidden layer), the matrix $G^{(r+1)}$ is the same for any input $x$. This means all inputs define the same output coefficients; equivalently, there is no output randomization for that layer. We use this fact to recover the signs of the neurons. The output coefficients are recovered via second derivatives, thus, this method resembles a second-order differential cryptanalysis.

Assume that the DNN has a single output and let $c_1,\dots,c_{d_r}$ be its output coefficients. Also, let $x$ be an input to the DNN and $y^{(i)}$ be the output of layer $i$ after the ReLUs, i.e., $y^{(i)} = F_{i}(x)$. The output of the DNN is given by
\begin{equation}
\label{eq:lastLayer_outputDNN}
    f(x) = c_1 y_1^{(r)} + \cdots + c_{d_r} y_{d_r}^{(r)} + b^{(r + 1)},
\end{equation}
where $ y_k^{(r)}$ is the $k$-th coordinate of $ y^{(r)}$ and $b^{(r+1)}$ is the bias of the output layer. 

With the recovered matrix $\hat{A}^{(r)}$, we can compute the value that neuron $k$ in layer $r$ takes before the ReLU, i.e., $e_k = \langle \hat{A}_{k}^{(r)}, y^{(r-1)} \rangle$, but we do not know its sign $s_k$. Let us consider both options. Exactly one of $e_k$ or $-e_k$ will be positive, and the other will be negative; the latter would be blocked by the ReLU. So, $\sigma(e_k, -e_k)$ is either $(e_k, 0)$ or $(0, -e_k)$, depending on whether $e_k > 0$ or $e_k < 0$, respectively. We know the value $e_k$, so we can compute $\sigma(e_k, -e_k)$. Let us write $(\hat{y}_{k+}^{(r)}, \hat{y}_{k-}^{(r)}) = \sigma(e_k, -e_k)$. Now, finding $s_k$ is equivalent to deciding whether the real value $y_k^{(r)}$ of the neuron after the ReLU is $\hat{y}_{k+}^{(r)}$ or $\hat{y}_{k-}^{(r)}$. The contribution to $f(x)$ in \autoref{eq:lastLayer_outputDNN} is $c_k \hat{y}_{k+}^{(r)}$ when $s_k = +1$, otherwise, it is $c_k \hat{y}_{k-}^{(r)}$. Then, that equation can be rewritten as
\[ f(x) = \sum_{k=1}^{d_r} c_k \left( s_k \hat{y}_{k+}^{(r)} + (1 - s_1) \hat{y}_{k-}^{(r)} \right) + b^{(r + 1)}. \]
So, we take random inputs $x$, construct a system of linear equations, and solve for the unknowns $s_k$ and $b^{(r + 1)}$. We must choose at least $d_r + 1$ random inputs for the system to have a unique solution. 

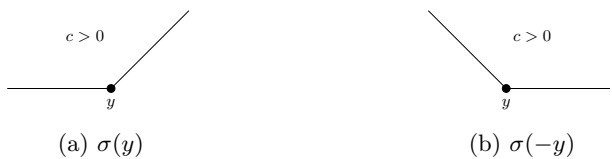
\begin{figure}[!ht]
	\centering
	\begin{subfigure}[t]{0.45\textwidth}
		\centering
		\resizebox{2.5cm}{!}{\begin{circuitikz}
\tikzstyle{every node}=[font=\small]

\draw (-2,0) -- (0,0) -- (1.5,1.5);
\draw[black,fill=black] (0,0) circle (0.75mm);
\node [font=\small] at (0,-0.3) {$y$};
\node [font=\small] at (-0.5,1) {$c > 0$};

\end{circuitikz}}
		\caption{$\sigma(y)$}
		\label{fig:reluSecondDerivative_ReLU(x)}
	\end{subfigure}
	\begin{subfigure}[t]{0.45\textwidth}
		\centering
		\resizebox{2.5cm}{!}{\begin{circuitikz}
\tikzstyle{every node}=[font=\small]

\draw (-1.5,1.5) -- (0,0) -- (2,0);
\draw[black,fill=black] (0,0) circle (0.75mm);
\node [font=\small] at (0,-0.3) {$y$};
\node [font=\small] at (0.5,1) {$c > 0$};

\end{circuitikz}}
		\caption{$\sigma(-y)$}
		\label{fig:reluSecondDerivative_ReLU(-x)}
	\end{subfigure}
	\caption{Second derivative $c = \frac{\sigma(y + \epsilon) - 2\sigma(y) + \sigma(y - \epsilon)}{\epsilon^2}$ for the ReLU function $\sigma$.}
	\label{fig:reluSecondDerivative}
\end{figure}

The output coefficients $c_k$ can be obtained through a second derivative; the latter represents the change in the slope of the function. \autoref{fig:reluSecondDerivative} depicts that the second derivative of the ReLU function $\sigma$ is the same at a critical point regardless of the sign of its input.
Let $x^*$ be a critical point for neuron $k$ in layer $i$ and $\Delta$ be a small-norm vector in the linear neighborhood of $x^*$. Then,
\[ f(x^* + \Delta) - 2f(x^*) + f(x^* - \Delta) = \pm \langle F_k^{(i)}, \Delta \rangle c_k, \]
where $F_k^{(i)}$ is the $k$-th row of the matrix $F^{(i)}$ defined by $x^*$ when collpasing layers $1$ to $i$. The sign above is $+$ when $x + \Delta$ activates the neuron, and it is $-$ otherwise. If $\Delta$ is parallel to $F_k^{(i)}$, $\langle F_k^{(i)}, \Delta \rangle > 0$ and the neuron is active with $x + \Delta$. Then, dividing the quantity above by $\langle F_k^{(i)}, \Delta \rangle$ yields the coefficient. When recovering the sign, we do not have access to the real $F^{(i)}$, but we know $\hat{F}^{(i)} = \hat{A}^{(i)} F^{(i-1)}$. To get the coefficient of neuron $k$, we choose $\Delta$ as $\hat{F}_k^{(i)}$ scaled to have a sufficiently small norm. Getting the output coefficient of a neuron can be done for any hidden layer. It is only in the last one that the coefficients remain constant for different points $x^*$.

\subsubsection{Oracle Calls/Time}

First, for each neuron, we find a critical point and compute the output coefficient. This requires $3$ oracle queries and linear algebra operations with time complexity $\mathcal O(d^3)$, where $d = \max(d_0, d_r)$. That is $3d_r$ queries and time complexity $\mathcal O(d_rd^3)$ for all $d_r$ neurons. Then, constructing the system of equations requires also $\mathcal O(d^3)$ operations (to compute the values $\hat{y}_{k-}^{(r)}$ and $\hat{y}_{k+}^{(r)}$ for $d_r + 1$ points) and $d_r + 1$ queries. Finally, solving the system of equations requires $\mathcal O(d_r^3)$ operations and no queries. In total, we require $4d_r + 1$ queries, and the time complexity is $\mathcal O(d_rd^3)$ operations.

\subsubsection{Limitations}

This technique requires the output coefficients to be constant. This only happens in the last hidden layer. Therefore, this method is applicable to that layer only.

\section{Practical Sign Recovery Attacks} \label{sec:experiments}

This section presents the experimental results of our proposed sign recovery techniques from~\autoref{sec:sign_recovery_techniques}. We first do a set of preliminary experiments on ``well-behaved'' unitary balanced neural networks of varying sizes (\autoref{sec:exp-random}), before we recover the real-world CIFAR10 network (\autoref{sec:exp-cifar10}). 

Our software implementation is available in\footnote{Anonymized for submission.}
\begin{center}
    \texttt{https://anonymous.4open.science/r/deti-C405}
\end{center}

We have executed the majority of our experiments on a DGX A100 server on a 40\,GiB GPU. However, our experiments are not GPU intensive and will run with similar runtimes on most personal computers.

\subsection{Implementation Caveats}
\label{sec:implementation}

\subsubsection{Infinite Numerical Precision}

The assumption of infinite numerical precision cannot be upheld practically, as conventional deep learning packages such as TensorFlow~\cite{tensorflow2015-whitepaper} only offer a maximal backend floating point precision of~64 bits. 
This practical limitation will influence all of our techniques: At infinite numerical precision, we could create an infinitesimal wiggle to be sure that we stay within the linear region of the neural network, but the limited floating point precision forces us to create a wiggle of sufficient magnitude, and this larger-magnitude wiggle, in turn, can accidentally toggle one of the other neurons. This requires us to perform a linearity check, which leads to additional oracle queries (see Appendix \ref{sec:findCriticalPoints}). The limited floating point precision also means that the small uncertainties in our recovered weights will propagate to larger errors in our prediction of neuron values at deeper layers. For both \SOE and the \LastLayer method, this means that the system of equations we recover can have slightly imprecise coefficients, and variables that should be exactly zero may instead resolve to a relatively small value. For all of the networks we studied, we found it sufficient to declare a small threshold for what should be considered a zero, but this line will get increasingly blurry with deeper networks.

\begin{figure}[!h]        
	\centering
	\begin{subfigure}[t]{.45\textwidth}
      \includegraphics[width=\textwidth]{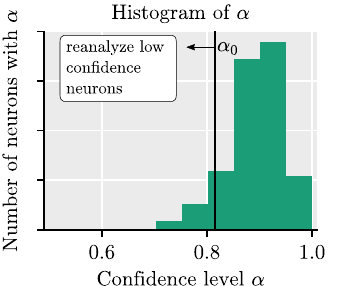}
      \caption{\NeuronWiggle: Adaptive $\alpha_0$.}
      \label{fig:adaptive-alpha}         
	\end{subfigure}
	\quad
	\begin{subfigure}[t]{.45\textwidth}
		\centering
  \includegraphics[width=\textwidth]{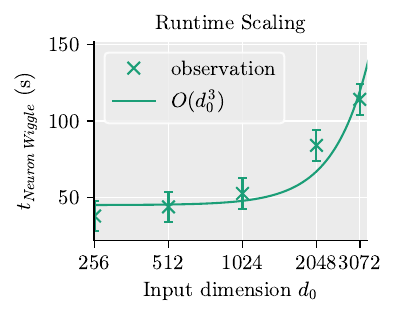}
  \caption{\NeuronWiggle: Runtime scaling.}
  \label{fig:runtimescaling}
	\end{subfigure}
	\caption{\subref{fig:adaptive-alpha}~illustrates that we reanalyze borderline neurons with low confidence level $\alpha<\alpha_0$ as explained in \nameref{sec:confidence-level}. 
    \subref{fig:runtimescaling}~shows the approximately cubic runtime scaling with the neural network input dimension investigated in \nameref{sec:runtime-cubic}.}\label{fig:two-experiments}
\end{figure}

\subsubsection{\NeuronWiggle: Confidence-level $\alpha$}\label{sec:confidence-level}
The neuron wiggle is a statistical method, and in all following experiments we use $s=200$ samples per neuron sign recovery. The sign recovery result for each neuron has a certain confidence level $\alpha\in (0.5,1.0]$. A confidence level of $\alpha=1.0$ ($\alpha\approx 0.5$) means absolute certainty (low certainty) in the recovered sign. 
The vast majority of signs is recovered \textbf{correctly} ({\color{cgrey}\cmark[]}), even if $\alpha\approx 0.5$. 
    However, a very small number of signs is recovered \textbf{incorrectly} ({\color{cgrey}\crossmark[]}). 
    We consider decisions with a confidence level below a cut-off value $\alpha_0$ to be \textbf{borderline} and re-analyze the neurons with $\alpha\leq \alpha_0$ after updating the signs for all neurons with high confidence level $\alpha> \alpha_0$. 
    The cut-off value $\alpha_0$ is chosen adaptively so that the neurons with the least-confident 10\%  of all sign-recoveries are reanalyzed (\autoref{fig:adaptive-alpha}). 
%

\subsection{Unitary Balanced Neural Networks}\label{sec:exp-random} 
We start with experiments on a set of ``well-behaved'' toy networks of varying sizes before tackling a ``real-world'' CIFAR10 network in the following section. 

\subsubsection{Methodology}
Training small networks is sometimes done on artificial datasets, such as random data in~\cite{carlini2020cryptanalytic}. 
Instead of training on random data, we purposefully create ``well-behaved'' networks with the following characteristics: 
The weights of each neuron are given by a unit vector chosen uniformly at random. The bias of each neuron is chosen such that it has a 50\% probability of being active. 
Using our unitary balanced networks, we avoid anomalies originating from the non-deterministic neural network training on random data. 


%

\subsubsection{Results on 784-128-1}
The largest network presented in \cite{carlini2020cryptanalytic} is the 784-128-1 network with around 100k parameters. Using \Freeze the time complexity for the sign recovery can be estimated as  $\mathcal O(d_i d^3)=128 \times 784^3 =2^{35} $. 
If \ExhaustiveS would have to be used, the time complexity for the sign recovery can be estimated as $\mathcal O(2^{d_i})=2^{128}$. 
While \cite{carlini2020cryptanalytic}  can still recover the 784-128-1 network in ``under an hour'', deeper networks with multiple hidden layers are untractable due to the sign recovery with exhaustive search. 
We can recover the signs of all neurons in the~128 neuron-wide hidden layer with either \SOE  
in about $(6.77\pm 0.04)$\,s, or alternatively, the \LastLayer technique 
in about $(18.6\pm0.05)$\,s. In our implementations of \SOE and \LastLayer, parallelization and further speed-up are trivially achievable. Note that although the complexities of the two methods are similar, the execution time of \LastLayer is mainly determined by the time it takes the algorithm to find the output coefficients, which is in turn dominated by the cost of finding critical points. Since these searches are independent, they can be done in parallel with a linear speedup factor. On the other hand, \SOE does not require critical points, which explains the slight gap in performance.

\subsubsection{Results on 100-200$^{(3)}$-10} 
This network is a larger version of the 10-20-20-1 network presented in \cite{carlini2020cryptanalytic}. As this network is expansive, the \Freeze technique and \SOE cannot be used, and before our work \ExhaustiveS with an estimated cost of $2^{200}$ would have had to be employed for sign recovery. We use the \NeuronWiggle technique in hidden layers~1 and~2, and the \LastLayer technique in hidden layer~3. 
In hidden layer~1 and~2 the \NeuronWiggle recovers all neuron signs correctly, even the ones with a low confidence level $\alpha\approx 0.5$. The (per-layer parallelizable) execution time per neuron is $(16.3\pm0.4)$\,s, respectively $(18.8\pm0.5)$\,s. 
The \LastLayer technique recovers hidden layer~3 in a total execution time of $(35.8\pm0.2)$\,s.

\subsubsection{Results on $d_0$-256$^{(8)}$-10}\label{sec:runtime-cubic}

Before moving on to our ``actual'' CIFAR10 network (3072-256$^{(8)}$-10), we investigate the runtime scaling on unitary balanced networks with varying input dimensions $d_0=\{256, 512, 1024, 2048, 3072\}$. 
According to our time order complexity estimation, we expect a cubic scaling of the runtime $\sim \mathcal O(s d_i d^3)$ with the input dimension $d=\max(d_0, d_i)$. 
For each network $d_0$-256$^{(8)}$-10, we recover a subset of neurons in hidden layer three. \autoref{fig:runtimescaling} shows an approximately cubic scaling with the shortest sign recovery time of 37\,s for $d_0=256$, and the longest sign recovery time of 114\,s for $d_0=3072$. 
\vspace{2em}

\subsection{CIFAR10 Neural Network}
\label{sec:exp-cifar10}

\subsubsection{The CIFAR10 Dataset} CIFAR10 is one of the typical benchmarking datasets in visual deep learning. It contains a balanced ten-class subset of the 80 Million Tiny Images \cite{torralba200880}. Each of the ten classes (e.g., airplane, cat, and frog) contains $32\times32$ pixel RGB images, totaling~50,000 training and~10,000 test images.
\subsubsection{Our CIFAR10 Model} \autoref{table:our-cifar-10-models} gives a detailed description of our CIFAR10 model. 
%
    \begin{table}[htb!]
    \caption{Description of our CIFAR10 model. Each image of the CIFAR10 dataset has $32\times32=1024$ pixels per RGB channel. Accordingly, our model has $3\times1024=3072$ input neurons. Further, it has 10 output neurons, one for each class in CIFAR10. We arbitrarily chose 256 neurons in each hidden layer. 
    Our 8-hidden-layer CIFAR10 model with around 1.25M parameters is the largest model we analyze in this manuscript.
    }
    \label{table:our-cifar-10-models}
    \begin{threeparttable}
    \footnotesize
    \renewcommand{\TPTminimum}{\linewidth}
    \makebox[\linewidth]{%
    \tabcolsep=0.11cm
    \setlength{\extrarowheight}{2pt}
    \begin{tabular}{lccccc}
     \bottomrule
    \specialrule{0.4pt}{\aboverulesep}{0pt}
     \thead[l]{model} & \thead[l]{acc. CIFAR10} &
      \thead[l]{\#(hidden layers)} & \thead[l]{\#(hidden neurons)} &
      \thead[l]{parameters} \\
      \Xhline{0.5pt}
 $3072-256^{(8)}-10$ &                0.5249 &                 8 &                     2048 &  1,249,802 \\
      \Xhline{0.8pt}
    \end{tabular}
     }
    \begin{tablenotes}[flushleft]\footnotesize\smallskip
    \item 
    Note: Column `model' shows the model configuration with the number of neurons in the (input layer, hidden layers$^{\rm \#(hidden\ layers)}$, output layer); 
    `acc. CIFAR10' shows the evaluation accuracy of the model on the 10,000 CIFAR10 test images;
    `\#(hidden layers)' shows the number of hidden layers; 
    `\#(hidden neurons)' shows the total number of hidden layer neurons;  
    `parameters' shows the total number of model weights and biases. 
   \par
   \end{tablenotes}
   \end{threeparttable}
   \end{table}
On the CIFAR10 dataset, we perform a standard rescaling of the pixel values from $0\ldots255$ to $0\ldots1$. For our model training, we choose typical settings (the optimizer is stochastic gradient descent with a momentum of~0.9; the loss is sparse categorical crossentropy; batch size 64), similar to~\cite{lin2015far}. 
Our eight-hidden-layer model shows CIFAR10 test accuracies within the expected range: 
Typical test accuracies for densely connected neural networks with pure ReLU activation functions are around $\approx 0.53$~\cite{lin2015far}. 
We note that better test accuracies are achieved using more advanced neural network architectures--beyond densely connected layers with pure ReLU activation functions. The current state of the art, Google Research's so-called vision transformer (ViT-H/14), achieves 0.995 test accuracy on CIFAR10~\cite[Table 2]{dosovitskiy2020image}. 

As mentioned previously, the \emph{space of control} is essential for the sign recovery, and here we provide the experimental analysis for our $3072-256^{(8)}-10$ CIFAR10 model. 
\autoref{table:positive-relus} shows that after a sharp initial drop, the space of control stabilizes in the deeper layers. This supports our intuition from \autoref{fig:rank-example} that a dramatic fall in the space of control is prevented by the affine transformations.

\begin{table}[htb!]
    \caption{The average number of ReLUs on their positive side in each hidden layer of our $3072-256^{(8)}-10$ CIFAR10 model for 10k random inputs, as well as the rank of a linearized representation $\mathcal M$ of the network up to the respective layer.}
    \label{table:positive-relus}
    \begin{threeparttable}
    \footnotesize
    \renewcommand{\TPTminimum}{\linewidth}
    \makebox[\linewidth]{%
    \tabcolsep=0.11cm
    \setlength{\extrarowheight}{2pt}
    \begin{tabular}{l|llllllll}
      \bottomrule
        \specialrule{0.4pt}{\aboverulesep}{0pt}
\thead[l]{$i$: Layer ID} &          1 &         2 &         3 &         4 &         5 &         6 &         7 &           8 \\
\thead[l]{\#(ON ReLUs)}   &  127$\pm$8 &  80$\pm$8 &  72$\pm$8 &  74$\pm$6 &  82$\pm$6 &  95$\pm$6 &  99$\pm$7 &  100$\pm$10 \\
\thead[l]{rank$(F^{(i)})$}    &  127$\pm$8 &  80$\pm$8 &  71$\pm$7 &  68$\pm$6 &  68$\pm$6 &  68$\pm$6 &  68$\pm$6 &    68$\pm$6 \\
    \Xhline{0.8pt}
    \end{tabular}
     }
   \end{threeparttable}
\end{table}

We analyzed our CIFAR10 network using our \SOE (hidden layers 1 and 2), \LastLayer (last hidden layer), and \NeuronWiggle (hidden layer three to penultimate) sign recovery techniques. 

\subsubsection{Results with SOE and Last Layer Technique} \autoref{table:cifar10-results-soe-last} shows the results of the sign recovery with the \SOE and \LastLayer techniques on our CIFAR10 model.
Since these methods are algebraic, all neuron signs can be recovered successfully. We collect a system of equations as described in \autoref{sec:soe} (\SOE) and \autoref{sec:last-layer} (\LastLayer). For \SOE, we collect one equation per hidden layer neuron (in total~256). Each equation $k=1,\dots,256$ contains the model output difference $f(x + \Delta_k) - f(x)$, and therefore, a total of $256+1$ oracle calls are necessary. Even in our non-optimized implementation, the runtimes are well below thirty seconds for the CIFAR10 model.
For the \LastLayer technique, we collect one equation per hidden layer neuron (in total~256) and one equation per output neuron bias (in total~10). 
%
\begin{table}[htb!]
    \caption{Results of the sign recovery with the \SOE and \LastLayer techniques on our $3072-256^{(8)}-10$ CIFAR10 model. 
    Each hidden layer contains 256 neurons. 
    All neuron signs can be recovered successfully (\cmark[]). 
    %
    }
    \label{table:cifar10-results-soe-last}
    \begin{threeparttable}
    \footnotesize
    \renewcommand{\TPTminimum}{\linewidth}
    \makebox[\linewidth]{%
    \tabcolsep=0.11cm
    \setlength{\extrarowheight}{2pt}
    \begin{tabular}{lgcc|lgcc}
     \bottomrule
    \specialrule{0.4pt}{\aboverulesep}{0pt} 
                      \multicolumn{4}{c}{\thead[c]{\SOE}} & \multicolumn{4}{c}{\thead[c]{\LastLayer}} \\ 
                      \cmidrule(lr){1-4} \cmidrule(lr){5-8} 
    layer ID & \cmark[] & queries & runtime & layer ID & \cmark[] & queries & runtime \\
    \Xhline{0.5pt}
     1,2 & \underline{256} & 256+1 & $(16\pm1)$\,s  & 8 & \underline{256} & 256+10 & $(189\pm40)$\,s \\ 
    \Xhline{0.8pt}
    \end{tabular}
     }
    \begin{tablenotes}[flushleft]\footnotesize\smallskip
    \item Column `queries' contains the number of oracle queries. 
    The runtime is the total execution time for recovering all neuron signs in seconds. For both, \SOE and \LastLayer, these execution times can be significantly reduced further by parallelizing the implementation.
   \par
   \end{tablenotes}
   \end{threeparttable}
   \end{table}

\subsubsection{Results with the Neuron Wiggle Technique} \autoref{table:cifar10-results} shows the results of the sign recovery with the \NeuronWiggle technique on our CIFAR10 model. 
\begin{table}[htb!]
    \caption{Results of the sign recovery with the \NeuronWiggle method on hidden layers $1,\dots,8$ of our CIFAR10 model 3072-256$^{(8)}$-10. 
    Each hidden layer contains 256 neurons. 
    The vast majority is evaluated \textbf{correctly} ({\color{cgrey}\cmark[]}), even at a low confidence level ({\color{cgrey}$\alpha\approx0.5$}). 
    \underline{256} highlights the cases where all neurons are analyzed correctly, even those with low confidence level $\alpha$. 
    Borderline neurons with a low confidence level ({\color{cgrey}$\alpha\leq\alpha_0$}) are re-evaluated.
    The column \textbf{fixable} contains the borderline neurons with wrongly recovered signs ({\color{cgrey}\crossmark[]}) and low confidence level. 
    Zero \textbf{incorrect} decisions ({\color{cgrey}\crossmark[]}) were made with a high confidence level ({\color{cgrey}$\alpha>\alpha_0$}). 
    %
    }
    \label{table:cifar10-results}
    \begin{threeparttable}
    \footnotesize
    \renewcommand{\TPTminimum}{\linewidth}
    \makebox[\linewidth]{%
    \tabcolsep=0.11cm
    \setlength{\extrarowheight}{2pt}
    \begin{tabular}{lc|bccgc}
     \bottomrule
    \specialrule{0.4pt}{\aboverulesep}{0pt}
                      & layerID 
                      & \thead[c]{$\alpha_0$\\(adaptive)}
                      & \thead[c]{\textbf{correct}
                        \\{\color{cgrey}($\alpha>0.5$) and \cmark[]}} 
					 & \thead[c]{\textbf{fixable}
                        \\{\color{cgrey}($\alpha\leq\alpha_0$) and \crossmark[]}} 
                      & \thead[c]{\textbf{incorrect}
                        \\\color{cgrey}{($\alpha>\alpha_0$) and \crossmark[]}} 
                      & \thead[c]{runtime} \\
    \Xhline{0.5pt}
& 1  &    0.79 &  255/256 & 1/256 & 0/256 &             1121 s \\
                  & 2  &    0.67 &  254/256 &  2/256 &        0/256 &               185 s \\
                  & 3  &    0.74 &  \underline{256}/256 &  \underline{0}/256  & 0/256 &               219 s \\
                  & 4  &    0.74 &  \underline{256}/256 & \underline{0}/256  &  0/256 &               295 s \\
                  & 5  &    0.74 &  \underline{256}/256 & \underline{0}/256  &  0/256 &               201 s \\
                  & 6  &    0.75 &  \underline{256}/256 & \underline{0}/256  &   0/256 &               269 s \\
                  & 7  &    0.70 &  253/256 &  3/256 &   0/256 &               234 s \\
                  & 8  &    0.77 &  252/256 &  4/256 &   0/256 &               384 s \\
    \Xhline{0.8pt}
    \end{tabular}
     }
    \begin{tablenotes}[flushleft]\footnotesize\smallskip
    \item 
    Note that our other two sign recovery techniques are faster for hidden layers 1, 2, and 8, and we only show the neuron wiggle timing for these layers for the sake of completeness. The column 'runtime' refers to the mean runtime per neuron in the target layer, where the signs of all the neurons in the same layer can be extracted in parallel on a multicore computer. 
   \par
   \end{tablenotes}
   \end{threeparttable}
   \end{table}
Each neuron sign was recovered using~200 critical points. We find criticals point for each target neuron by starting from a randomly chosen input image from the CIFAR10 dataset.
As detailed in \autoref{sec:confidence-level}, the neuron wiggle technique is statistical. 
We reanalyze borderline neurons with a low confidence level $\alpha\leq\alpha_0$. \\
Detailed neuron-by-neuron results for hidden layer 7 (of 8) are shown in \autoref{sec:detailed-cifar10}. The mean sign recovery time per neuron in the relevant layers with layer IDs~$2,\dots,8$ is $\bar t_{\rm total}=(234\pm44)$\,seconds. 
Note that the sign recovery of the~256 neurons within the same hidden layer can be parallelized. Therefore, a parallelized implementation of our sign recovery algorithm on a suitable 256-core computer would take around 4 minutes per layer and a total of about 32 minutes for the 8-hidden layer DNN. 

Compared to our preliminary experiments on the unitary balanced networks, the sign recovery time for the comparable-size CIFAR10 model is almost twice as long. The detailed analysis in \autoref{sec:detailed-cifar10} shows that the increased execution time is due to finding critical points: in the actual CIFAR10 network, it takes around 110\,s, while it only took 10\,s in the unitary balanced network. Notable differences to the unitary balanced network are a smaller \emph{space of control} and the existence of \emph{always-off} or \emph{almost-always-off} neurons, which makes finding critical points more challenging.

\section{Conclusions}
\label{sec:conc}

In this paper, we presented the first polynomial-time algorithm for extracting the parameters of ReLU-based DNNs from their black-box implementation. We also demonstrated the practical efficiency of our sign recovery by applying it to a deep network for classifying CIFAR10 images. We used SOE ($i=1,2$, $(16\pm1)$\,s per layer), Neuron Wiggling ($i=3\ldots7$, $(234\pm44)$\,s per layer) and Last Hidden Layer ($i=8$, $(189\pm40)$\,s). 
The total required time is about~30~minutes. 
We also demonstrated its applicability to several layer deep, expanding networks whose width was much larger than the number of inputs (where all the previous techniques based on solving systems of linear equations failed). \\
Among the many problems left open are:
\begin{enumerate}
\item Developing countermeasures and countercountermeasures for our attack.
\item Dealing with the case in which only the class decisions are provided rather than the exact logits.
\item Dealing with more modern machine learning architectures such as transformers.
\item Dealing with discrete domains such as texts in LLMs where derivatives are not well defined.
\end{enumerate}

%
%
\bibliographystyle{alpha} 
\bibliography{bibliography}

\clearpage
\newpage

\appendix

\section{The Expected Signal-to-Noise Ratio of Neuron Wiggle in Unitary Balanced Networks}
\label{sec:neuronWiggleUnitaryBalancedNetworks}

In this appendix, we perform an approximate back-of-the-envelope analysis of the relative sizes of the signal and the noise in the neural wiggling technique described in \autoref{sec:neuron-wiggle} for unitary balanced networks. Recall that, in this model, the weights of all neurons are unit vectors chosen uniformly at random. Here, we assume that unit vectors of dimension $n$ have the form $(\pm\frac{1}{\sqrt{n}},\dots,\pm\frac{1}{\sqrt{n}})$. The input wiggle $\delta$ for the $i$-th layer has tiny norm $\varepsilon$, but to simplify our notation, we will assume that it has norm $1$ since $\varepsilon$ is just a common multiplicative factor in both the signal and the noise.

Let us first analyze the situation where the number of degrees of freedom at the input to layer $i$ is also $n$ (i.e., the attacker has full control of this input wiggle). Let $u$ be the weight vector of the target neuron and $v$ the weight vector of a different neuron in the same layer. Then, $\norm{\dotp{u}{\delta}} = 1$ (since $u$ and $\delta$ are unit vectors in the same direction) and $\norm{\dotp{v}{\delta}}$ is expected to be about $ \frac{1}{\sqrt{n}}$. To see the latter, write a unit vector as $\frac{1}{\sqrt{n}}(\pm1,\dots,\pm1)$, thus $\norm{\dotp{v}{\delta}}$ is $1/n$ times the sum of $n$ entries each one of which is either $+1$ or $-1$. This sum can be viewed as a simple random walk whose expected translation distance after n steps is $\sqrt{n}$. This yields the estimate that the size of the dot product of two random unit vectors in $n$ dimensional space is about $\norm{\dotp{v}{\delta}} = \frac{1}{\sqrt{n}}$.

Without loss of generality, assume that the target neuron is the first neuron in the layer. Let $e_k$ be the difference for neuron $k$ given the wiggle $\delta$. Then, the vector of differences for the target layer is
\begin{align*}
    (e_1, e_2, \dots, e_n) &= (e_1, 0, \dots, 0) + (0, e_2, \dots, e_n) \\
                           &= \underbrace{(1, 0, \dots, 0)}_{e_t} + \underbrace{\left(0, \pm\frac{1}{\sqrt{n}}, \dots, \pm\frac{1}{\sqrt{n}}\right)}_{e_o},
\end{align*}
where the second equality holds by the discussion above. The vector $e_t$ represents the change in the value for the target neuron, and $e_o$ the change in the value for all other neurons. The norm of $e_t$ is $1$. Notice that $e_o$ is very close to being a unit vector (only the first entry is zero); hence, its norm is close to $1$. Then, the norm of the change in value of the target neuron is similar to the norm of the changes in value of all the other neurons combined, so the size of the signal is comparable to the size of the noise at the input to the $i$-layer ReLU's, making the signal easily detectable.

When the attacker has only $d$ degrees of freedom at the input to layer $i$, the signal diminishes while the noise remains the same. To model this effect, assume that in the unit vector $(\pm\frac{1}{\sqrt{n}},\dots,\pm\frac{1}{\sqrt{n}})$ which represents the desired wiggle, the attacker can optimally choose only the first $d$ signs, while the remaining $n-d$ signs are randomly chosen. This will reduce the norm of the signal $\norm{\dotp{u}{\delta}}$ from $1$ to about $\sqrt{d/n}$ while keeping the norm of the noise essentially unchanged (since it is still determined by the dot products of independently chosen random unit vectors). This represents a very gradual deterioration in the signal-to-noise ratio as $d$ decreases; for example, when the attacker has only $d=n/2$ degrees of freedom, the signal-to-noise ratio is only reduced by a multiplicative factor of about $\sqrt{1/2}=0.7$. While the signal is expected to be a bit smaller than the noise, it will still be easy to detect after enough samples are taken. This explains the success of our experiments on expanding networks with three consecutive layers of $200$ neurons while the network had only $100$ inputs.

\section{Finding Critical Points and Determining Linearity}
\label{sec:findCriticalPoints}

We are interested in finding the points where a piecewise linear function changes its slope. Particularly, when the function is given by the output of a DNN, each of these points corresponds to a critical point for a neuron. Here, we show a method to find these points, and it builds on the ideas presented by Carlini et al.~\cite{carlini2020cryptanalytic} and Jagielski et al.~\cite{jagielski2020high}. This method improves the robustness for both finding critical points and detecting false critical points.

Consider one output of the DNN. Given two inputs $X_1, X_2 \in \mathbb{R}^{d_0}$, we can think of them as being the boundaries of a closed interval in $\mathbb{R}$. For example, each point $\delta \in [0, 1] \subset \mathbb{R}$ corresponds to $X_1 + \delta (X_2 - X_1)$. Thus, we will consider the DNN to be a function whose inputs are real values in an interval $[x_{\alpha}, x_{\beta}] \subset \mathbb{R}$.

The main idea is to detect whether there is exactly one critical point in $[x_{\alpha}, x_{\beta}]$. To do this, compute the point $x^*$ where the line passing through $x_{\alpha}$ intersects the line passing through $x_{\beta}$. Compute the expected value of $f$ at that point, denoted by $\widehat{f}(x^*)$. Then, $x^*$ is a critical point if $\widehat{f}(x^*) = f(x^*)$. If multiple critical points are detected, partition the interval and search recursively within $[x_{\alpha}, x_o]$ and $[x_o, x_{\beta}]$, where $x_o = (x_{\alpha} + x_{\beta})/2$.

\begin{algorithm}
\caption{\texttt{FindCriticalPoints}}
\label{alg:Find_critical_points}
\begin{algorithmic}[1]
    \REQUIRE{$x_{\alpha}, x_{\beta}, \varepsilon \in \mathbb{R}$}
    \ENSURE{Set of critical points}

    \STATE{$x_o \leftarrow \frac{x_{\alpha} + x_{\beta}}{2}$}
    \STATE{$y_{\alpha} \leftarrow f(x_{\alpha})$, $y_{\beta} \leftarrow f(x_{\beta})$, $y_{o} \leftarrow f(x_{o})$}
    \IF{$y_{o} = \frac{y_{\alpha} + y_{\beta}}{2}$}
        \STATE{$m_{\alpha} \leftarrow \frac{f(x_{\alpha} + \varepsilon) - y_{\alpha}}{\varepsilon}$, $m_{\beta} \leftarrow \frac{y_{\beta} - f(x_{\beta} - \varepsilon)}{\varepsilon}$}
        \IF{$m_{\alpha} = m_{\beta}$}
            \STATE{$m_{\alpha, o} \leftarrow \frac{y_o - f(x_o - \varepsilon)}{\varepsilon}, m_{\beta, o} \leftarrow \frac{f(x_o + \varepsilon) - y_o}{\varepsilon}$}
            \IF{$m_{\alpha, o} = m_{\alpha}$ \AND $m_{\beta, o} = m_{\beta}$}
                \RETURN{$\emptyset$}
            \ENDIF
        \ENDIF
    \ENDIF
    \STATE{$x^* = x_{\alpha} + \dfrac{y_{\beta} - y_{\alpha} - m_{\beta}(x_{\beta} - x_{\alpha})}{m_{\alpha} - m_{\beta}}$}
    \STATE{$\widehat{f}(x^*) = y_{\alpha} + m_{\alpha}\dfrac{y_{\beta} - y_{\alpha} - m_{\beta}(x_{\beta} - x_{\alpha})}{m_{\alpha} - m_{\beta}}$}
    \IF{$f(x^*) = \widehat{f}(x^*)$ \AND $f'_{-}(x^*) \neq f'_{+}(x^*)$ \AND $x^* \in [x_{\alpha}, x_{\beta}]$}
        \RETURN{$\{x^*\}$}
    \ENDIF
    \RETURN{\texttt{FindCriticalPoints($x_{\alpha}, x_0, \varepsilon$)} $\cup$ \texttt{FindCriticalPoints($x_0, x_{\beta}, \varepsilon$)}}
\end{algorithmic}
\end{algorithm}

\begin{figure}
    \centering
    \begin{subfigure}[b]{0.32\textwidth}
        \centering
        \resizebox{4cm}{!}{\includegraphics[width=\textwidth]{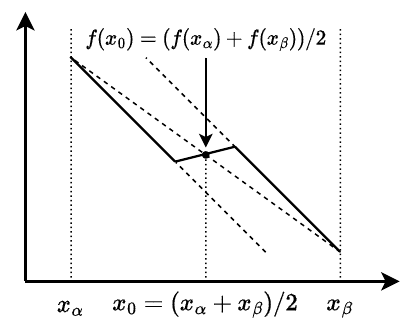}}
        \caption{Not a critical point.}
        \label{fig:CP_Parallel}
    \end{subfigure}
    \hfill
    \begin{subfigure}[b]{0.32\textwidth}
        \centering
        \resizebox{4cm}{!}{\includegraphics[width=\textwidth]{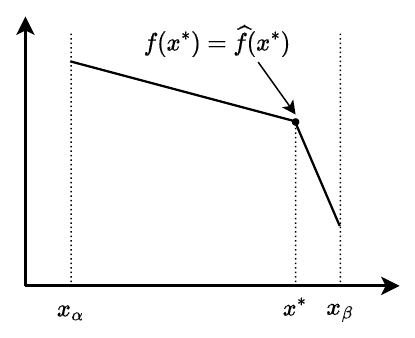}}
        \caption{Critical point.}
        \label{fig:CP_InterPt_Correct}
    \end{subfigure}
    \hfill
    \begin{subfigure}[b]{0.32\textwidth}
        \centering
        \resizebox{4cm}{!}{\includegraphics[width=\textwidth]{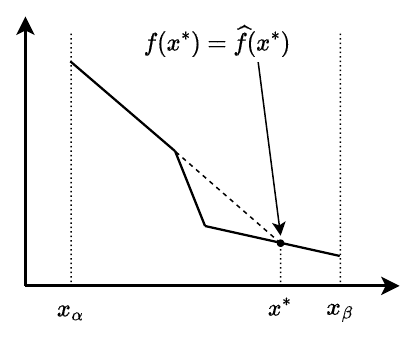}}
        \caption{Not a critical point.}
        \label{fig:CP_InterPt_Incorrect01}
    \end{subfigure}
    ~
    \begin{subfigure}[b]{0.32\textwidth}
        \centering
        \resizebox{4cm}{!}{\includegraphics[width=\textwidth]{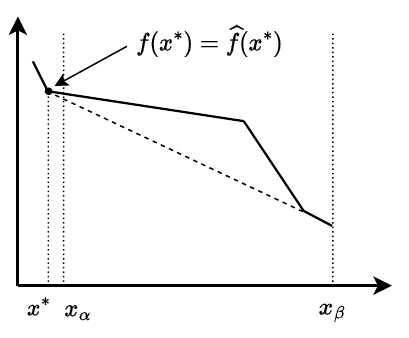}}
        \caption{Critical point outside $[x_{\alpha}, x_{\beta}]$.}
        \label{fig:CP_InterPt_Incorrect02}
    \end{subfigure}
    \hfill
    \begin{subfigure}[b]{0.32\textwidth}
        \centering
        \resizebox{4cm}{!}{\includegraphics[width=\textwidth]{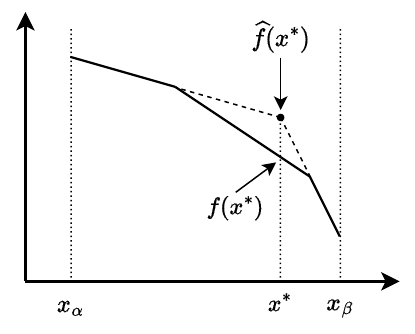}}
        \caption{Not a critical point.}
        \label{fig:CP_InterPt_NoCP}
    \end{subfigure}
    \caption{Finding critical points.}
    \label{fig:FindingCP}
\end{figure}

The details are in Algorithm~\ref{alg:Find_critical_points}. The first part is to check whether $f$ is linear in $[x_{\alpha}, x_{\beta}]$ (lines 1-11). To do this, we first check if $f(x_o) = (f(x_{\alpha}) + f(x_{\beta}))/2$. Checking only this condition fails in the particular case depicted in Figure~\ref{fig:CP_Parallel}. Thus, we next compare the slopes of the lines passing through $x_{\alpha}$ and $x_{\beta}$. If the slopes are different, there are multiple critical points. Equality means that they are either the same line or parallel lines (as in Figure~\ref{fig:CP_Parallel}). The former case implies no critical points (i.e., linearity) while the latter implies multiple critical points within the interval. We distinguish between these cases with the slope of the lines passing through the left and right side of $x_o$. If $f$ is not linear in $[x_{\alpha}, x_{\beta}]$, we compute a candidate $x^*$ and check whether it is the only critical point (lines 12-16). This occurs when (i) $f(x^*) = \widehat{f}(x^*)$, (ii) the left and right derivatives of $f$ at $x^*$ are different and (iii) $x^* \in [x_{\alpha}, x_{\beta}]$. Condition (i) is sufficient if there is exactly one critical point within $[x_{\alpha}, x_{\beta}]$; see Figure~\ref{fig:CP_InterPt_Correct}. When the interval contains more than one critical point, conditions (ii) and (iii) avoid erroneously reporting $x^*$ as one of them in certain cases; see Figures~\ref{fig:CP_InterPt_Incorrect01} and \ref{fig:CP_InterPt_Incorrect02}, respectively. Figure \ref{fig:CP_InterPt_NoCP} shows the scenario when $\widehat{f}(x^*) \neq f(x^*)$. If multiple critical points are detected, we partition the interval and continue the search.

Our sign recovery techniques require that we evaluate the DNN with input differences in the linear neighborhood of a given point. With infinite precision, we can simply choose an arbitrarily small input difference. However, our implementation has a limit on the precision of arithmetic operations. As mentioned in \autoref{sec:implementation}, this forces us to check for linearity when we evaluate small differences in the input to ensure that we do not toggle any neuron. We do this in the same way as explained above.

\newpage
\clearpage
\section{Detailed Results for CIFAR10}\label{sec:detailed-cifar10}

\begin{table}[htb!]
    \captionsetup{font=footnotesize}
    \caption{Detailed neuron-by-neuron results for hidden layer 7 (of 8) in the CIFAR10 model~3072-256$^{(8)}$-10. 
    We highlight the three cases where the sign recovery gave an incorrect result (\crossmark[]). 
    Note the low confidence level $\alpha$ in these cases. 
    We also highlight the longest observed recovery time $t_{\rm total}=684$\,s. The mean recovery time is $\bar t_{\rm total}=(234\pm49)$\,seconds. The median recovery time is $\tilde t_{\rm total}=220$\,s, and 75\% of all neurons can be recovered within 227\,s.
    }
    \label{table:cifar10-results-details}
    \begin{threeparttable}
    \footnotesize
    \renewcommand{\TPTminimum}{\linewidth}
    \makebox[\linewidth]{%
    \tabcolsep=0.11cm
    \setlength{\extrarowheight}{2pt}
    \resizebox{7.5cm}{!}{
    \begin{tabular}{lcllccccc}
     \bottomrule
    \specialrule{0.4pt}{\aboverulesep}{0pt}
 neuronID & real sign & $s_-$ & $s_+$ & $\alpha$ &      correct & $t_{\rm crit.}$ & $t_{\rm wiggle}$ & $t_{\rm total}$ \\
\Xhline{0.5pt}
 neuron 0 &        - &        -:165 &        +:35 &   0.82 & \cmark[] &       135 &       84 &                 221 \\
 neuron 1 &        + &          -:0 &       +:200 &   1.00 & \cmark[] &       130 &       83 &                 213 \\
 neuron 2 &        - &        -:198 &         +:2 &   0.99 & \cmark[] &       138 &       83 &                 224 \\
 neuron 3 &        - &        -:176 &        +:24 &   0.88 & \cmark[] &       131 &       83 &                 216 \\
 neuron 4 &        - &        -:161 &        +:39 &   0.81 & \cmark[] &       132 &       83 &                 217 \\
 neuron 5 &        + &          -:0 &       +:200 &   1.00 & \cmark[] &       128 &       83 &                 213 \\
 neuron 6 &        - &        -:193 &         +:7 &   0.96 & \cmark[] &       130 &       84 &                 216 \\
 neuron 7 &        + &         -:14 &       +:186 &   0.93 & \cmark[] &       148 &       83 &                 232 \\
 neuron 8 &        - &        -:200 &         +:0 &   1.00 & \cmark[] &       128 &       83 &                 213 \\
 neuron 9 &        - &        -:125 &        +:75 &   0.62 & \cmark[] &       174 &       83 &                 258 \\
neuron 10 &        - &        -:200 &         +:0 &   1.00 & \cmark[] &       129 &       83 &                 213 \\
neuron 11 &        - &        -:195 &         +:5 &   0.97 & \cmark[] &       137 &       83 &                 223 \\
neuron 12 &        - &        -:199 &         +:1 &   0.99 & \cmark[] &       128 &       83 &                 214 \\
neuron 13 &        - &        -:194 &         +:6 &   0.97 & \cmark[] &       133 &       83 &                 218 \\
neuron 14 &        + &          -:1 &       +:199 &   0.99 & \cmark[] &       132 &       83 &                 216 \\
neuron 15 &        + &          -:3 &       +:197 &   0.98 & \cmark[] &       139 &       83 &                 227 \\
 $\ldots$ &  $\ldots$ & $\ldots$     & $\ldots$    & $\ldots$&  $\ldots$   &  $\ldots$ & $\ldots$ &       $\ldots$      \\
neuron 108 &        - &        -:199 &         +:1 &   0.99 &     \cmark[] &       133 &       83 &                 217 \\
neuron 109 &        + &        -:105 &        +:95 &   0.53 & \crossmark[] &       496 &       83 &                 581 \\
neuron 110 &        + &         -:51 &       +:149 &   0.74 &     \cmark[] &       144 &       83 &                 230 \\
 $\ldots$ &  $\ldots$ & $\ldots$     & $\ldots$    & $\ldots$&  $\ldots$   &  $\ldots$ & $\ldots$ &       $\ldots$      \\
neuron 108 &        - &        -:199 &         +:1 &   0.99 &     \cmark[] &       133 &       83 &                 217 \\
neuron 109 &        + &        -:105 &        +:95 &   0.53 & \crossmark[] &       496 &       83 &                 581 \\
neuron 110 &        + &         -:51 &       +:149 &   0.74 &     \cmark[] &       144 &       83 &                 230 \\
 $\ldots$ &  $\ldots$ & $\ldots$     & $\ldots$    & $\ldots$&  $\ldots$   &  $\ldots$ & $\ldots$ &       $\ldots$      \\
neuron 116 &        + &          -:2 &       +:198 &   0.99 &     \cmark[] &       152 &       83 &                 239 \\
neuron 117 &        - &         -:87 &       +:113 &   0.56 & \crossmark[] &       302 &       83 &                 390 \\
neuron 118 &        + &          -:1 &       +:199 &   0.99 &     \cmark[] &       134 &       83 &                 219 \\
 $\ldots$ &  $\ldots$ & $\ldots$     & $\ldots$    & $\ldots$&  $\ldots$   &  $\ldots$ & $\ldots$ &       $\ldots$      \\
neuron 245 &        + &          -:0 &       +:200 &   1.00 & \cmark[] &       132 &       84 &                 218 \\
neuron 246 &        - &        -:177 &        +:23 &   0.89 & \cmark[] &       597 &       83 &                 684 \\
neuron 247 &        + &          -:6 &       +:194 &   0.97 & \cmark[] &       135 &       84 &                 222 \\
  $\ldots$ &  $\ldots$ & $\ldots$     & $\ldots$    & $\ldots$&  $\ldots$   &  $\ldots$ & $\ldots$ &       $\ldots$      \\
neuron 251 &        - &        -:184 &        +:16 &   0.92 & \cmark[] &       153 &       83 &                 237 \\
neuron 252 &        - &        -:200 &         +:0 &   1.00 & \cmark[] &       135 &       84 &                 222 \\
neuron 253 &        - &        -:200 &         +:0 &   1.00 & \cmark[] &       132 &       83 &                 217 \\
neuron 254 &        - &        -:200 &         +:0 &   1.00 & \cmark[] &       132 &       83 &                 218 \\
neuron 255 &        - &        -:195 &         +:5 &   0.97 & \cmark[] &       132 &       84 &                 218 \\
    \Xhline{0.8pt}
    \end{tabular}
    }
        }
    \begin{tablenotes}[flushleft]\footnotesize\smallskip
    \item 
    Note: For each neuron, 200 critical points were analyzed for the sign recovery. $s_\pm$ negative and positive sign votes. Confidence level $\alpha$. Time to find critical points $t_{\rm crit.}$, and the execution time for the wiggle sign recovery itself $t_{\rm wiggle}$. $t_{\rm total}=t_{\rm crit.}+t_{\rm wiggle}$.
   \par
   \end{tablenotes}
   \end{threeparttable}
   \end{table}

   \newpage



\end{document}